%% file: main.tex
\documentclass[10pt,twocolumn,letterpaper]{article}

\usepackage[pagenumbers]{cvpr}

\usepackage{graphicx}
\usepackage{amsmath}
\usepackage{amssymb}
\usepackage{booktabs}
\usepackage{microtype}
\usepackage{makecell}
\usepackage{float}
\usepackage{caption}
\usepackage{subcaption}
\usepackage{lineno}
\usepackage{epsfig}
\usepackage{comment}
\usepackage{multirow}
\usepackage{amsfonts}
\usepackage{xcolor}
\usepackage{adjustbox}
\usepackage{textcomp}
\usepackage{gensymb}
\usepackage{url}
\usepackage{setspace}
\usepackage{bm}
\usepackage{wrapfig}
\usepackage[accsupp]{axessibility}  
\usepackage[ruled,linesnumbered,noend]{algorithm2e}
\def \HiLi{\leavevmode\rlap{\hbox to \hsize{\color{yellow!50}\leaders\hrule height .8\baselineskip depth .5ex\hfill}}}

\usepackage[pagebackref,breaklinks,colorlinks]{hyperref}

\usepackage{enumitem}

\usepackage[capitalize]{cleveref}
\crefname{section}{Sec.}{Secs.}
\Crefname{section}{Section}{Sections}
\Crefname{table}{Table}{Tables}
\crefname{table}{Tab.}{Tabs.}

\begin{document}

\title{Towards Compositional Adversarial Robustness: \\
Generalizing Adversarial Training to Composite Semantic Perturbations}
\author{Lei Hsiung$^{1,4}$,~~Yun-Yun Tsai$^2$,~~Pin-Yu Chen$^{3}$,~~Tsung-Yi Ho$^{1,4}$ \\
$^1$National Tsing Hua University, $^2$Columbia University, $^3$IBM Research, \\
$^4$The Chinese University of Hong Kong}

\twocolumn[{
\renewcommand\twocolumn[1][]{#1}
\maketitle
\vspace{-10mm}
\begin{center}
\textbf{\url{https://hsiung.cc/CARBEN/}}
\end{center}
}]

\begin{abstract}
Model robustness against adversarial examples of single perturbation type such as the $\ell_{p}$-norm has been widely studied, yet its generalization to more realistic scenarios involving multiple semantic perturbations and their composition remains largely unexplored. In this paper, we first propose a novel method for generating composite adversarial examples. Our method can find the optimal attack composition by utilizing component-wise projected gradient descent and automatic attack-order scheduling. We then propose \textbf{generalized adversarial training} (\textbf{GAT}) to extend model robustness from $\ell_{p}$-ball to composite semantic perturbations, such as the combination of Hue, Saturation, Brightness, Contrast, and Rotation. Results obtained using ImageNet and CIFAR-10 datasets indicate that GAT can be robust not only to all the tested types of a single attack, but also to any combination of such attacks. GAT also outperforms baseline $\ell_{\infty}$-norm bounded adversarial training approaches by a significant margin.
\end{abstract}

\input{preprint/docs/1_introduction}
\input{preprint/docs/2_relatedwork}
\input{preprint/docs/3_method}
\input{preprint/docs/4_experiment}
\input{preprint/docs/5_conclusion}

\section{Acknowledgement}
Part of this work was done during L. Hsiung's visit to IBM Thomas J. Watson Research Center. The work of L. Hsiung and T.-Y. Ho was supported by the Ministry of Science and Technology, Taiwan (MOST 111-2218-E-005-006-MBK). We thank National Center for High-performance Computing (NCHC) in Taiwan for providing computational and storage resources.

{\small
\bibliographystyle{ieee_fullname}
\bibliography{main}
}

\onecolumn
\appendix
\setcounter{table}{0}
\setcounter{figure}{0}
\renewcommand{\thetable}{A\arabic{table}}
\renewcommand{\thefigure}{A\arabic{figure}}
\input{preprint/docs/6_appendix}

\end{document}

%% file: preprint/docs/1_introduction.tex
\section{Introduction}\label{sec:introduction}
Deep neural networks have shown remarkable success in a wide variety of machine learning (ML) applications, ranging from biometric authentication (e.g., facial image recognition), medical diagnosis (e.g., CT lung cancer detection) to autonomous driving systems (traffic sign classification), etc. However, while these models can achieve outstanding performance on benign data points, recent research has shown that state-of-the-art models can be easily fooled by malicious data points crafted intentionally with adversarial perturbations \cite{szegedy2013intriguing}.

To date, the most effective defense mechanism is to incorporate adversarial examples during model training, known as adversarial training (AT) \cite{madry2017towards, zhang2019theoretically}.
Nonetheless, current adversarial training approaches primarily only consider a single perturbation type (or threat model) quantified in a specific distance metric (e.g., $\ell_{p}$-ball). In this regard, the lack of exploration of the compositional adversarial robustness against a combination of several threat models could lead to impractical conclusions and undesirable bias in robustness evaluation. For example, a model that is robust to perturbations within $\ell_{p}$-ball does not imply it can simultaneously  be robust to other realistic semantic perturbations (e.g., hue, saturation, rotation, brightness, and contrast).

To tackle this issue, in this paper, we propose  \textbf{generalized adversarial training (GAT)}, which can harden against a wide range of threat models, from single $\ell_{\infty}$-norm or semantic perturbation to a combination of them. Notably, extending standard adversarial training to composite adversarial perturbations is a challenging and non-trivial task, as each perturbation type is sequentially applied, and thus the attack order will affect the effectiveness of the composite adversarial example. To bridge this gap, we propose an efficient attack order scheduling algorithm to learn the optimal ordering of various perturbation types, which will then be incorporated into the GAT framework.

\begin{figure*}[t]
  \centering
  \begin{subfigure}[b]{\textwidth}
      \centering
      \includegraphics[width=.96\textwidth, trim=1.4cm 10.3cm 1.27cm 1.2cm, clip]{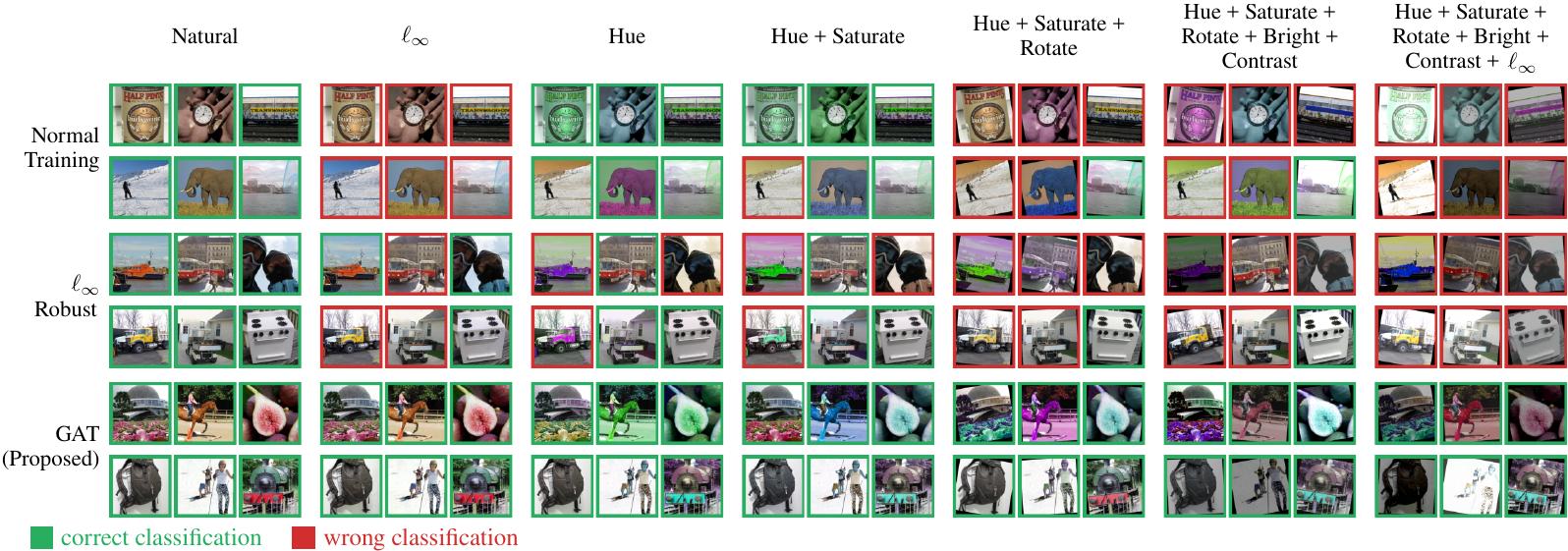}
      \caption{}
      \label{fig:composite_attack_exp}
  \end{subfigure}
  \hfill
  \begin{subfigure}[b]{\textwidth}
      \centering
      \includegraphics[width=.96\textwidth, trim=1.9cm 0.5cm 2cm 0.4cm, clip]{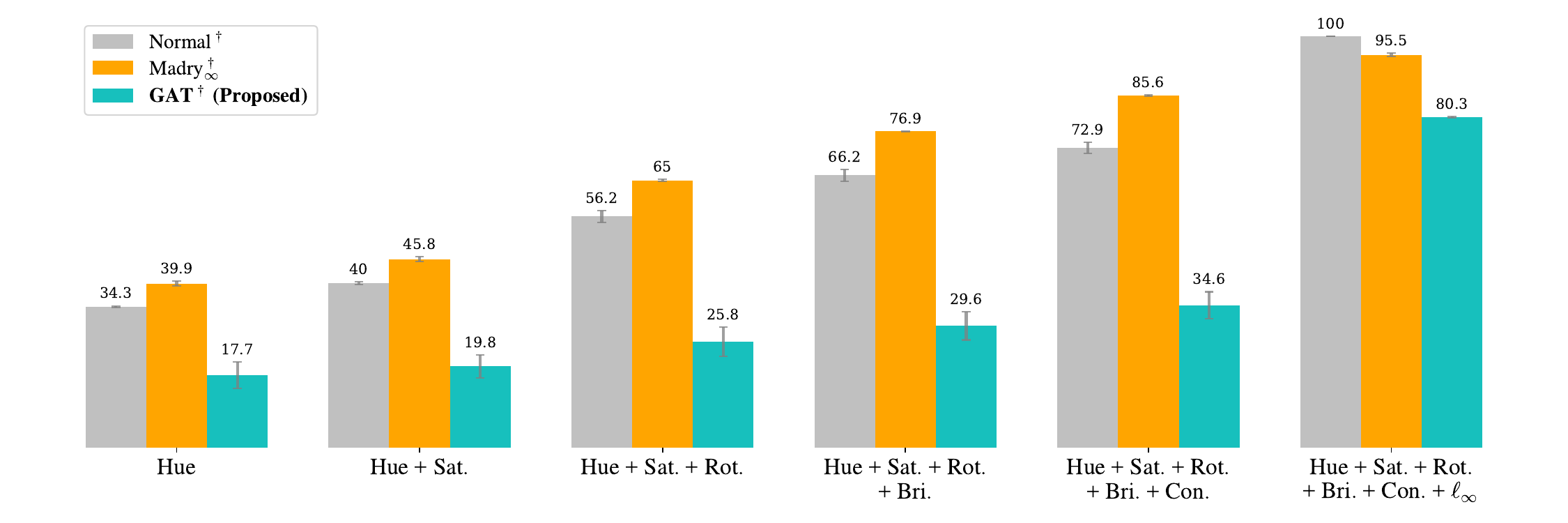}
      \caption{}
      \label{fig:composite_attack_eval}
  \end{subfigure}
\vspace{-5mm}
  \caption{(a) \textbf{Qualitative study:} illustration of some perturbed examples generated by different attack combinations and their predictions by different ResNet50 models \cite{He2015DeepRL} on ImageNet, including standard training, Madry's $\ell_{\infty}$ robust training \cite{madry2017towards} and our proposed \textbf{GAT}. The results show that our proposed GAT can maintain robust accuracy under a variety of composite adversarial attacks, even with the increasing number of attacks. (b) \textbf{Quantitative study:} the attack success rate (ASR, \%) of the above-mentioned models under multiple composite attacks (a higher ASR means less robust) on all correctly classified test samples for each model.
  The corresponding robust accuracy (RA) is listed in Table \ref{tab:ra_with_diff_attack_num}.}  \vspace{-5mm}
  \label{fig:demo_fig}
\end{figure*}

Different from existing works, this paper aims to address the following fundamental questions:
(a) How to generalize adversarial training from a single threat model to multiple? (b) How to optimize the perturbation order from a set of semantic and $\ell_{p}$-norm perturbations? (c) Can GAT outperform other adversarial training baselines against composite perturbations? 

Our main contributions in this paper provide affirmative answers to the questions:
\vspace{-2mm}
\begin{enumerate}[leftmargin=*]
    \item We propose composite adversarial attack (CAA), a novel and unified approach to generate adversarial examples across from multiple perturbation types with attack-order-scheduling, including semantic perturbations (\textit{Hue, Saturation, Contrast, Brightness and Contrast}) and $\ell_{p}$-norm space. To the best of our knowledge, this paper is the first work that leverages a scheduling algorithm for finding the optimal attack order in composite adversarial attacks.  
    
    \vspace{-2mm}\item Building upon our composite adversarial attack framework, we propose generalized adversarial training (\textbf{GAT}) toward achieving compositional adversarial robustness, which enables the training of neural networks robust to composite adversarial attacks.
    \vspace{-2mm}\item For the attack part, our proposed composite adversarial attack exhibits a high attack success rate (ASR) against standard or $\ell_{\infty}$-norm robust models. Moreover, our method with learned attack order significantly outperforms random attack ordering, giving an average $9$\% and $7$\% increase in ASR on CIFAR-10 and ImageNet.
    \vspace{-2mm}\item For the defense part, comparing our GAT to other adversarial training baselines \cite{madry2017towards, zhang2019theoretically, laidlaw2021perceptual, zhang2020fat, wu2020adversarial, wong2020fast},
    the results show the robust accuracy of GAT outperforms them by average 30\% $\sim$ 60\% on semantic attacks and 15\% $\sim$ 22\% on full attacks.
\end{enumerate}
\vspace{-2mm}

To further motivate the effectiveness of our proposed GAT framework, Fig. \ref{fig:demo_fig} compares the performance of different models under selected attacks, ranging from a single threat to composite threats. The models include standard training, $\ell_\infty$-robust training, and our proposed GAT. The results show the limitation of $\ell_\infty$-robust model \cite{madry2017towards}, which is robust against the same-type adversarial attack, but becomes fragile against semantic adversarial attacks and their composition. Our proposed GAT addresses this limitation by providing a novel training approach that is robust to any combination of multiple and adversarial threats.

%% file: preprint/docs/2_relatedwork.tex
\section{Related Work}\label{sec:relatedwork}
\paragraph{Adversarial Semantic Perturbations.}
Adversarial machine learning research has largely focused on generating examples that can deceive models into making incorrect predictions \cite{biggio2018wild}. One widely studied class of attacks involves $\ell_{p}$-norm adversarial perturbations \cite{goodfellow2014explaining, carlini2017towards, chen2017ead, croce2020reliable}. However, natural transformations such as changes in geometry, color, and brightness can also cause adversarial vulnerabilities, leading to what are known as semantic perturbations \cite{hosseini2018semantic, joshi2019semantic, shamsabadi2020colorfool, wang2021demiguise, wang2020generating, bhattad2019unrestricted, kang2019testing, qiu2020semanticadv}. Unlike $\ell_{p}$-norm perturbations, semantic perturbations often result in adversarial examples that look natural and are semantically similar to the original image but have significant differences in the $\ell_{p}$-norm perspective.

Hosseini and Poovendran \cite{hosseini2018semantic} demonstrated that randomly shifting the Hue and Saturation components in the Hue-Saturation-Value (HSV) color space of images can significantly reduce the accuracy of a neural network by up to 88\%. Bhattad et al. \cite{bhattad2019unrestricted} proposed similar attacks, including colorization and texture transfer attacks, which can perturb a grayscale image with natural colorization or infuse the texture of one image into another. Prior work on geometric transformations has targeted rotation transformations. For example, Xiao et al. \cite{xiao2018spatially} used coordinate-wise optimization at each pixel, which can be computationally expensive. Engstrom et al. \cite{engstrom2019exploring} proposed a simple way of parametrizing a set of tunable parameters for spatial transformations. Dunn et al. \cite{dunn2020evaluating} exploited context-sensitive changes to features from the input and perturbed images with the corresponding feature map interpolation. Mohapatra et al. \cite{Mohapatra_2020_CVPR} studied certified robustness against semantic perturbations, but they did not discuss adversarial training.

\vspace{-4mm}
\paragraph{Composite Adversarial Perturbations.}
Previous literature has inspired researchers to explore different metrics \cite{wong2019wasserstein, laidlaw2021perceptual} and the combination of various adversarial threats \cite{laidlaw2019functional, bhattad2019unrestricted, jordan2019quantifying, yuan2022adaptive} to harden adversarial examples. These works have expanded the perturbation space of an image and have successfully increased the misclassification rate of neural networks. For instance, Laidlaw and Feizi \cite{laidlaw2019functional} propose the ReColorAdv attack, which combines multi-functional threats to perturb every input pixel and also includes additional $\ell_{p}$-norm threat. Mao et al. \cite{mao2020composite} utilized genetic algorithms to search for the best combination of multiple attacks, which were found to be stronger than a single attack. However, their approach only considered searching the order of attack combination in specific norm spaces (i.e., $\ell_{2}$, $\ell_{\infty}$, and corruption semantic space) and could not handle all attacks simultaneously. On the other hand, Yuan et al. \cite{yuan2022adaptive} have incorporated different image transformation operations to improve the transferability of adversarial examples.

Regarding measuring model robustness, Kang et al. \cite{kang2019testing} proposed utilizing ensemble unforeseen attacks from broader threat models, including JPEG, Fog, Snow, Gabor, etc. They consider the worst-case scenario over all attacks and attempt to improve model performance against these unforeseen adversarial threats. Prior works have shown that combining different types of adversarial threats can result in more robust adversarial examples. Our work builds upon these ideas and proposes a method for scheduling multiple attack types to generate composite adversarial perturbations.

\vspace{-3mm}
\paragraph{Adversarial Training (AT).}
AT is a widely adopted method for improving model robustness against adversarial attacks \cite{kurakin2016adversarial_ICLR, madry2017towards, zhang2019theoretically, Zhou2021TowardsDA}. One of the pioneering works in this field is by Madry et al. \cite{madry2017towards}, who proposed to minimize the worst-case loss in a region around the input. Zhang et al. \cite{zhang2019theoretically} further improved AT by considering both natural and adversarial inputs in computing the loss, along with a parameter $\beta$ to define the ratio of them, resulting in a smoother robust decision boundary. Laidlaw et al. \cite{laidlaw2021perceptual} expanded adversarial attacks from single to multiple threat models by using neural perceptual distance measurement to generalize adversarial training with perceptual adversarial examples. Recently, Mao et al. \cite{mao2022towards} proposed to combine robust components as building blocks of vision transformers, leading to a state-of-the-art robust vision transformer. AT with adversarial transformations is also done in \cite{stutz2020confidence, engstrom2019exploring}. 

While most previous works on AT have focused on improving model robustness against a single threat model, as shown in Fig. \ref{fig:demo_fig}, a model that is robust against $\ell_{\infty}$-norm perturbations may still have low robustness against composite semantic attacks or other $\ell_{q}$ threats ($p \neq q$) \cite{sharma2017attacking}. This has led researchers to consider multiple-norm adversarial training \cite{tramer2019adversarial,wang2019towards, pmlr-v119-maini20a}, which yields models that are simultaneously robust against multiple $\ell_{p}$-norm attacks. Tramer et al. \cite{tramer2019adversarial} have considered alternately optimizing perturbation types given a fixed attack order, but the search for the strongest possible attack order is left out of their discussion. Also, the considered perturbations are simultaneously added to the same data sample rather than sequentially.

In contrast to prior arts, this paper offers a novel approach to improving model robustness against multiple adversarial threats. Our attack takes into account the efficient attack order scheduling and extends beyond the $\ell_p$-norm attacks by incorporating various semantic perturbations, which can result in more robust adversarial examples. By incorporating the composite adversarial examples, our defense mechanism can significantly improve the robustness of the model. Overall, our work represents an important step towards creating more robust deep learning models that can defend against a wide range of adversarial attacks.

%% file: preprint/docs/3_method.tex
\section{Methodologies (CAA \& GAT)}\label{sec:method}
In this section, we first propose the composite adversarial attack (CAA) framework (Fig. \ref{fig:caa_implementation}), and elucidate the details of our attack order scheduling algorithm. We then adopt the CAA into adversarial training, which is called generalized adversarial training (GAT).

\subsection{Composite Attack Formulation}\label{subsec:composite_attack_formulation}
\begin{figure}
    \centering
    \includegraphics[width=\linewidth]{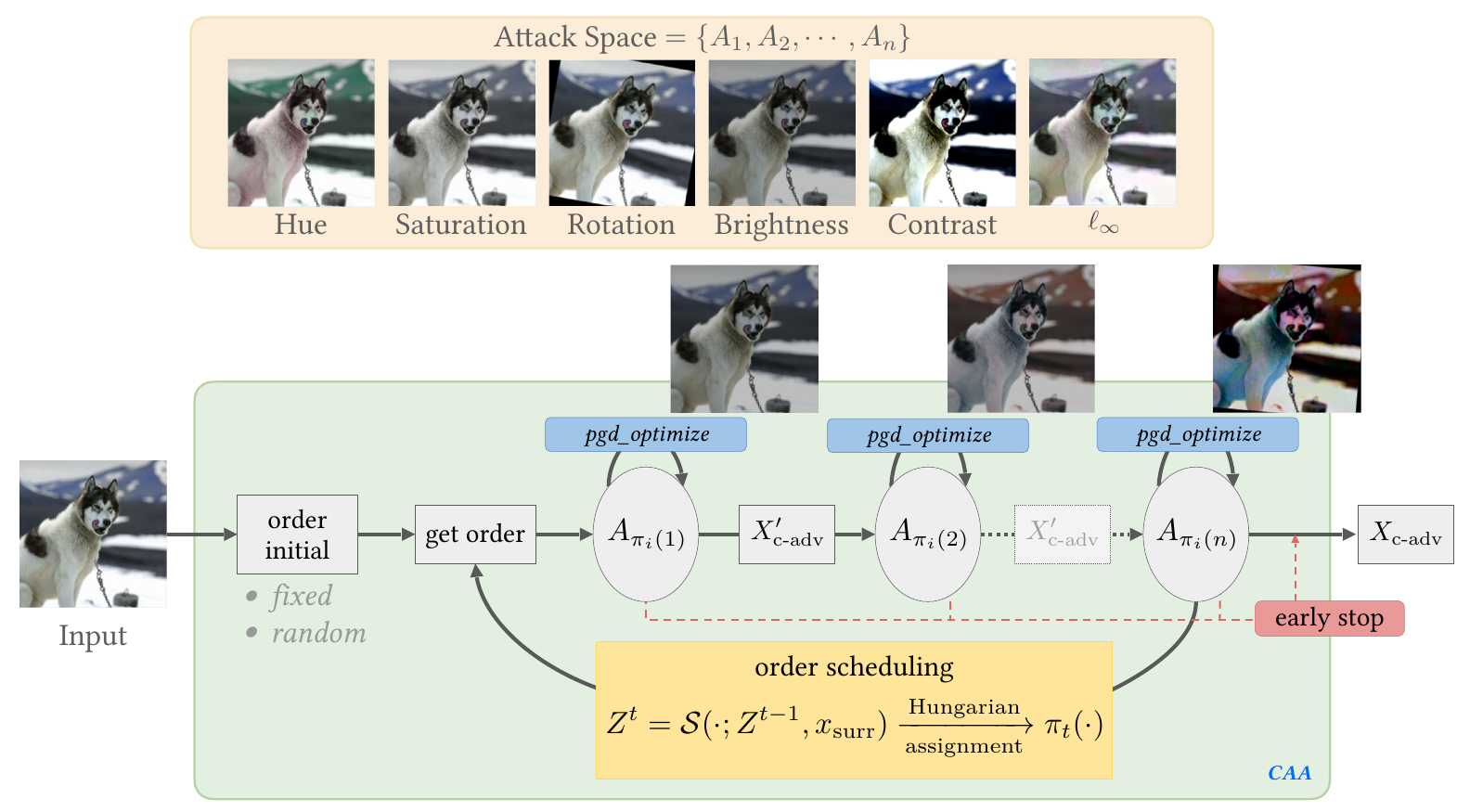}
    \vspace{-7mm}
    \caption{A pipeline of the proposed \textit{composite adversarial attack} method with the ability to dynamically optimize the attack order and harden adversarial examples.}
    \label{fig:caa_implementation}
    \vspace{-5mm}
\end{figure}

\paragraph{Order Scheduling.} 
Let $\mathcal{F}: \mathcal{X}\rightarrow\mathbb{R}^d$ be an image classifier that takes image $x\in\mathcal{X}$ as input and generates a $d$-dimensional prediction scores (e.g., softmax outputs) for $d$ classes, and let $\Omega=\{A_1,\ldots, A_n\}$ denote an attack set that contains $n$ attack types. For each attack $A_k$, we define a corresponding perturbation interval (boundary) $\epsilon_k = [\alpha_k,\beta_k]$ to govern the attack power of $A_k$. We then denote the corresponding perturbation intervals of $\Omega$ as $E=\{\epsilon_{k}|k\in\{1,\ldots,n\}\}$.

In CAA, we optimize not only the power of each attack component in $\Omega$, but also the attack order applied to the image $x$. 
That is, consider $\mathcal{I}_n=\{i\}_{i=1}^{n}$, we can use an assignment function $\pi_{i}: \mathcal{I}_n\to\mathcal{I}_n$ to determine the attack order to be used under the $i$-th schedule. As shown in Fig. \ref{fig:caa_implementation}, after $i$-th scheduling, a composite adversarial example $x_\text{c-adv}$ can be formulated as:
\vspace{-1mm}
\begin{align*}
x_\text{c-adv} = A_{\pi_{i}(n)}(A_{\pi_{i}(n-1)}(\cdots A_{\pi_{i}(1)}(x))).
\end{align*}

\vspace{-1mm}\noindent Noted that input $x$ would be perturbed in the order of: $A_{\pi_{i}(1)} \to A_{\pi_{i}(2)} \to \cdots \to A_{\pi_{i}(n)}$. For each attack operation $A_{k} \in \Omega$, an input $x$ would be transformed to a perturbed sample with a specific perturbation level $\delta_{k}$, where $\delta_{k}\in\epsilon_{k}$ would be optimized via projected gradient descent, maximizing the classification error (e.g., cross-entropy loss $\mathcal{L}$). Therefore, the operation of $A_k(x;\delta_k)$ could be expressed as optimizing $\delta_k$, that is:
\vspace{-1mm}
\begin{align}
\label{eqn:adv}
   \mathop{\arg\max}\limits_{ \delta_k\in\epsilon_k} 
   \mathcal{L} (\mathcal{F}(A_{k}(x;\delta_{k})),y),
\end{align}

\vspace{-1mm}\noindent where $y$ is the ground-truth label of $x$. We named it component-wise PGD (Comp-PGD) and will explain more details in Sec. \ref{subsec:comp_pgd}.

Since the assignment function $\pi_{i}(\cdot)$ is essentially a permutation matrix (or Birkhoff polytope), we can optimize it by treating it as a (relaxed) \textit{scheduling matrix} $Z^{i}$, where $Z^{i}=\big[\mathbf{z}_{1},\ldots,\mathbf{z}_{n}\big]^\top$ is also a doubly stochastic matrix, i.e. $\mathbf{z}_j\in\mathbb{R}^{n}$, $\sum_{i}{z_{ij}}=\sum_{j}{z_{ij}}=1$, $\forall~i,j\in\{1,\ldots,n\}$.
Furthermore, we can utilize the Hungarian algorithm \cite{kuhn1955hungarian, Munkres1957AlgorithmsFT} to obtain an optimal attack order assignment.

In sum, we formalize CAA's attack order auto-scheduling as a constrained optimization problem, where the attack order having maximum classification error can be obtained by solving:
\vspace{-1mm}
\begin{align}
    \label{eqn:advorder}
    \mathop{\max}\limits_{ \pi}
    \mathcal{L} (\mathcal{F}(A_{\pi(n)}  (\cdots A_{\pi(1)}(x;\delta_{\pi(1)}) \cdots;\delta_{\pi(n)}) ),y)\text{.}
\end{align}

\paragraph{The Surrogate Image for Scheduling Optimization.}
Since $x_\text{c-adv}$ contains merely one attack perturbation at each iteration, using it alone is challenging to  optimize the likelihood of other attacks in the relaxed scheduling matrix. To manage this issue, we adopt a surrogate composite adversarial image $x_\text{surr}$ to relax the restriction and compute the loss for updating the scheduling matrix $Z$, i.e. by weighting each type of attack perturbation with its corresponding probability at each iteration. Therefore, we could optimize the scheduling matrix $Z$ via maximizing the corresponding loss $\mathcal{L}(\mathcal{F}(x_\text{surr}),y)$.
Given the attack pool $\Omega$ of $n$ attacks, the surrogate image would be computed for $n$ iterations. For each iteration $i$, the surrogate image is defined as:
\vspace{-1mm}
\begin{align}
{x_\text{surr}^i}=\sum_{j=1}^{n}{z_{ij}}\cdot{A_{j}(x_\text{surr}^{i-1};\delta_{j})})\text{, }\forall i\in{\{1,\ldots,n\}}\text{,}
\end{align}

\vspace{-2mm}\noindent and $x_\text{surr}^0 = x$. Let $\mathbf{A}^{\top}=\big(A_{1},\ldots,A_{n}\big)$ denotes a vector of all attack types in $\Omega$. Consequently, after $n$ iterations, the resulting surrogate image $x_\text{surr}^n$ can be formulated into the following compositional form:
\vspace{-1mm}
\begin{equation}
\label{eqn:surrogate_image}
\begin{aligned}
x_\text{surr}^n &= \mathbf{z}_n^{\top}\mathbf{A}(\cdots(\mathbf{z}_{2}^{\top}\mathbf{A}(\mathbf{z}_{1}^{\top}\mathbf{A}(x)))) \\
& = \mathbf{z}_n^{\top}\mathbf{A}(\cdots(\mathbf{z}_{2}^{\top}\mathbf{A}(\sum_{j=1}^{n}{z_{1j}}\cdot{A_{j}(x;\delta_{j})}))) \\
& = \mathbf{z}_n^{\top}\mathbf{A}(\cdots(\mathbf{z}_{2}^{\top}\mathbf{A}(x_\text{surr}^1))) \\[3pt]
& = \mathbf{z}_n^{\top}\mathbf{A}(\cdots(x_\text{surr}^2)).
\end{aligned}
\end{equation}

\vspace{-3mm}
\paragraph{How to Learn Optimal Attack Order?}
Learning an optimal attack order expressed by the scheduling matrix $Z^\star$ is originally a combinatorial optimization problem to solve the best column and row permutation of a scheduling matrix. Sinkhorn and Knopp proved that any positive square matrix could be turned into a doubly stochastic matrix by alternately normalizing the rows and columns of it \cite{sinkhorn_1966}. Furthermore, Mena et al. theoretically showed how to extend the Sinkhorn normalization to learn and determine the optimal permutation matrix \cite{Mena2018Sinkhorn}. Similarly, in our problem, optimizing the attack order over a doubly stochastic matrix $Z$ can be cast as a maximization problem, where the feasible solution set is convex. With the surrogate composite adversarial example $x_\text{surr}$, the updating process of the scheduling matrix $Z^t$ for iteration $t$ can be formulated as:
\vspace{-1mm}
\begin{align}
\label{eqn:dsm_update}
    Z^{t} &= \mathcal{S}\big(\exp(Z^{t-1}+\frac{\partial \mathcal{L}(\mathcal{F}(x_\text{surr}),y)}{\partial Z^{t-1}})\big) \text{,}
\end{align}

\vspace{-1mm}
\noindent where $\mathcal{S}$ (Sinkhorn normalization) can be done in a limited number of iterations \cite{sinkhorn1967concerning}. Here, we fixed the iteration as 20 steps. After deriving an updated scheduling matrix, we utilize the Hungarian assignment algorithm to obtain the updated order assignment function $\pi_{t}(\cdot)$, as shown in Eq. \ref{eqn:hungarian}:
\vspace{-1mm}
\begin{align}
\label{eqn:hungarian}
    \pi_{t}(j) := \arg \max \mathbf{z}_{j} \text{, } \forall j \in \{1,\ldots,n\}.
\end{align}

\subsection{The Component-wise PGD (Comp-PGD)}\label{subsec:comp_pgd}
Upon addressing the attack scheduling issue, we now move forward to elucidate the design of adversarial perturbation in each attack type (component) of our composite adversarial attacks. For most of the semantic perturbations, their parameters are of continuous value. Therefore, we propose to search the parameters of semantic attacks by gradient descent algorithm within each continuous semantic space. In particular, we showed how to optimize the parameters in the following five different semantic perturbations, including (i) hue, (ii) saturation, (iii) brightness, (iv) contrast, and (v) rotation.  We extend the iterative gradient sign method \cite{Kurakin2016Adversarial} to optimize our semantic perturbations for $T$ iterations, which is defined as:
\vspace{-1mm}
\begin{align}
\label{eqn:attack_pgd}
    \delta_{k}^{t+1} = \text{clip}_{\epsilon_{k}}  \big( \delta_{k}^{t} + \alpha\cdot\text{sign}(\nabla_{\delta_{k}^{t}}\mathcal{L}(\mathcal{F}(A_{k}(x;\delta_{k}^{t})),y))\big)\text{,}
\end{align}

\vspace{-1mm}
\noindent where $t$ denotes the iteration index, $\alpha$ is the step size of each iteration, $\nabla_{\delta_{k}^{t}}\mathcal{L}(\cdot)$ is the gradient of a loss function $\mathcal{L}$ with respect to the perturbation variable $\delta_{k}^{t}$.
Let $\epsilon_k=[\alpha_k,\beta_k]$, we denote the element-wise clipping operation $\text{clip}_{\epsilon_k}(z)$ as: 
\vspace{-1mm}
\begin{align*}
\text{clip}_{\epsilon_k}(z) =
\text{clip}_{[\alpha_k,\beta_k]}(z) =
\left\{ \begin{array}{rl}
\alpha_{k} & \mbox{if } z < \alpha_{k} \text{,} \\ 
z & \mbox{if } \alpha_{k} \leq z \leq \beta_{k} \text{,}  \\
\beta_{k} & \mbox{if } \beta_{k} < z \text{.}
\end{array}\right.
\end{align*}

Next, we elucidate each semantic attack. The concrete examples of each of them are shown in Appendix \ref{sec:appendix_attack_levels} and the loss trace analysis of Comp-PGD is shown in Appendix \ref{sec:appendix_loss_landscape}.
\vspace{-2mm}
\paragraph{Hue.} The hue value is defined on a color wheel in HSV color space, ranging from $0$ to $2\pi$. In hue attack ($A_{H}$), we define the perturbation interval of hue as $\epsilon_{H}=[\alpha_{H},\beta_{H}]$, $-\pi\leq\alpha_{H}\leq\beta_{H}\leq\pi$.
Let $x_{H}=\text{Hue}(x)$ denote the hue value of an image $x$, the variation of hue value at step $t$ is $\delta_{H}^{t}$, and the initial variance $\delta_{H}^{0}$ is chosen from $\epsilon_{H}$ uniformly. Then $\delta_{H}^t$ can be updated iteratively via Eq. \ref{eqn:attack_pgd}, and the hue value of the perturbed image $x_\text{c-adv}^t=A_{H}(X;\delta_{H}^t)$ is:
\vspace{-1mm}
\begin{align*}
    x_{H}^t = \text{Hue}(x_\text{c-adv}^t) = \text{clip}_{[0,2\pi]}(x_{H}+\delta_{H}^t)\text{.}
\end{align*}

\vspace{-5mm}\paragraph{Saturation.}
Similar to hue value, saturation value determines the colorfulness of an image ranging from $0$ to $1$. Let $x_{S}=\text{Sat}(x)$ denote the saturation value of an image $x$. If $x_{S}\to0$, the image becomes more colorless, resulting in a gray-scale image if $x_{S}=0$. The perturbation interval of saturation is defined as $\epsilon_{S}=[\alpha_{S},\beta_{S}]$, $0\leq\alpha_{S}\leq\beta_{S}<\infty$.
Let the perturbation factor of saturation value at step $t$ is $\delta_{S}^{t}$, and the initial factor $\delta_{S}^{0}$ is chosen from $\epsilon_{S}$ uniformly. The saturation attack is to update the perturbation factor $\delta_{S}$ via Eq. \ref{eqn:attack_pgd}, and the saturation value of the perturbed image $x_\text{c-adv}^t=A_{S}(X;\delta_{S}^t)$ is:
\vspace{-1mm}
\begin{align*}
    x_{S}^t = \text{Sat}(x_\text{c-adv}^t) = \text{clip}_{[0,1]}( x_{S}\cdot\delta_{S}^{t})\text{.}
\end{align*}
\vspace{-8mm}\paragraph{Brightness and Contrast.}
Unlike hue and saturation, these values are defined on RGB color space (pixel space), and they determine the lightness, darkness, and brightness differences of images. In our implementation, we convert the images from $[0,255]$ scale to $[0,1]$. The perturbation interval of brightness and contrast is defined as $\epsilon_{B}=[\alpha_{B},\beta_{B}]$, $-1\leq\alpha_{B}\leq\beta_{B}\leq1$ and $\epsilon_{C}=[\alpha_{C},\beta_{C}]$, $0\leq\alpha_{C}\leq\beta_{C}<\infty$, respectively; the same, the initial perturbation $\delta_{B}^{0}$ and $\delta_{C}^{0}$ are chosen from $\epsilon_{B}$ and $\epsilon_{C}$ uniformly, and can update via Eq. \ref{eqn:attack_pgd}. The perturbed image $x_\text{c-adv}^t$ under the brightness attack ($A_{B}$) and contrast attack ($A_{C}$) can be formulated as:
\vspace{-1mm}
\begin{align*}
    x_\text{c-adv}^{t} = \text{clip}_{[0,1]}(x+\delta_{B}^{t}) \text{~~and~~}
    x_\text{c-adv}^{t} = \text{clip}_{[0,1]}(x\cdot\delta_{C}^{t})\text{.}
\end{align*}
\vspace{-6mm}\paragraph{Rotation.}
This transformation aims to find a rotation angle such that the rotated image has a maximum loss. The rotation implementation is constructed by \cite{riba2020kornia}. Given a square image $x$, let $(i,j)$ denotes pixel position and $(c,c)$ denotes the center position of $x$. Then the position $(i^{\prime},j^{\prime})$ rotated by $\theta$ degree from $(i,j)$ can be formulated as:
\vspace{1mm}
\begin{align*}
    \begin{bmatrix}
        {i}^{\prime}\\
        {j}^{\prime}\\
    \end{bmatrix}
    & = 
    \begin{bmatrix}
        \cos\theta\cdot{i} + \sin\theta\cdot{j} + (1-\cos\theta) \cdot c -  \sin\theta\cdot c \\ 
        -\sin\theta\cdot{i} + \cos\theta\cdot{j} + \sin\theta\cdot c +  (1-\cos\theta)\cdot c  \\
    \end{bmatrix}\text{.}
\end{align*}

\noindent Here, we define the perturbation interval of rotation degree $\epsilon_{R}=[\alpha_{R}\degree,\beta_{R}\degree]$, $\alpha_{R}\leq\beta_{R}$, $\alpha_{R},\beta_{R}\in\mathbb{R}$. The perturbation degree at step $t$ is $\delta_{R}^{t}$, and the initial degree $\delta_{R}^{0}$ is chosen from $\epsilon_{R}$ uniformly.
Similarly, like the previous attack, the perturbation $\delta_{R}$ will be updated via Eq. \ref{eqn:attack_pgd}.

\subsection{Generalized Adversarial Training (GAT)}
To harden the classifier against composite perturbations, we generalize the standard adversarial training approach with our proposed composite adversarial attack from Section \ref{subsec:composite_attack_formulation}. 
Our goal is to train a robust model $\mathcal{F}(\cdot)$ over a data distribution $(x,y) \sim \mathcal{D}$, and make it robust against composite perturbations in the perturbation boundary $E$.
Existing adversarial training objectives such as the $\min$-$\max$ loss \cite{madry2017towards} or TRADES loss \cite{zhang2019theoretically} can be utilized in GAT. Here we use $\min$-$\max$ training loss (Madry's loss) for illustration.
The inner maximization in Eq. \ref{eqn:adv_training} is to generate $x_{\text{c-adv}}$ optimized using CAA within boundary $E$, and the outer minimization is for optimizing the model parameters $\theta_{\mathcal{F}}$.
\vspace{-1mm}
\begin{align}
    \label{eqn:adv_training}
    \min_{\theta_{\mathcal{F}}}\mathop{\mathbb{E}}_{(x,y)\sim\mathcal{D}}
    \bigg[
    \max_{x_{\text{c-adv}}\in \mathcal{B}(x; \Omega; E)} \mathcal{L}(\mathcal{F}(x_{\text{c-adv}}),y)\bigg]\text{.}
\end{align}

For completeness, in Appendix \ref{sec:appendix_gat_algo} we summarize the flow of our proposed composite adversarial attacks with order scheduling and attack component optimization. In addition, the ablation study showing order-scheduling and Comp-PGD are essential can be found in Appendix \ref{sec:appendix_random_training}.

%% file: preprint/docs/4_experiment.tex
\section{Experiments}\label{sec:experiment}
In this section, we first elucidate the experimental settings and then present the performance evaluation and analysis against multiple composite attacks on two datasets: CIFAR-10 \cite{Krizhevsky09learningmultiple} and ImageNet \cite{ILSVRC15}. 
Additional experimental results and implementation details are shown in Appendix \ref{sec:appendix_additional_exp}.

\subsection{Experiment Setups}\label{sub_sec:experiment_setup}
\paragraph{Datasets.}
We evaluated GAT on two different datasets: CIFAR-10 \cite{Krizhevsky09learningmultiple} and ImageNet \cite{ILSVRC15}. CIFAR-10 consists of 60000 32*32 images, with 6000 images per class. There are 50000 training images and 10000 test images. ImageNet is a benchmark in image classification and object detection with 10 million images, including 1000 classes.

\vspace{-4mm}
\paragraph{Attack Composition.}
There are many feasible combinations of threats can be utilized in the evaluation; we discuss two attack combinations here, \textit{semantic attacks} and \textit{full attacks}, with two scheduling strategies. Semantic attacks consist of a combination of five semantic perturbations, including \textit{Hue}, \textit{Saturation}, \textit{Rotation}, \textit{Brightness} and \textit{Contrast} attacks.
For full attacks, one can generate examples with \textit{all five semantic attacks} and $\ell_{\infty}$ \textit{attack}. We consider different order scheduling strategies: \textit{scheduled} and \textit{random}. 
That is, we can either schedule the order by the aforementioned scheduling algorithm in Sec. \ref{subsec:composite_attack_formulation}, or randomly shuffle an attack order when launching attacks for generating the corresponding composite adversarial examples. Furthermore, we also present the results of a variety of attack compositions for analysis (see Appendix \ref{sec:appendix_sen_analysis_order})
and discuss the difference between separately/jointly optimizing the attack parameters in Appendix \ref{sec:appendix_sep_opt}.

\vspace{-4mm}
\paragraph{Comparative Training Methods.}\label{subsec:model_detail}
We compare our GAT with several baseline adversarial training models on both datasets using two different model backbones: ResNet50 \cite{He2015DeepRL} and WideResNet34 \cite{Zagoruyko16WRN}. The comparative methods are summarized in \textbf{Baseline Model Details} below.
For GAT, we train our models via finetuning on the $\ell_{\infty}$-robust pretrained model for both CIFAR-10 and ImageNet and use the min-max loss in Eq. \ref{eqn:adv_training} \cite{madry2017towards}. Two ordering modes were adopted in GAT: random order (\textit{GAT-f}) and scheduled order (\textit{GAT-fs}). We also found that training from scratch using GAT is unstable due to the consideration of multiple perturbation threats (see Appendix \ref{sec:appendix_implementation_details}).

\vspace{-4mm}
\paragraph{Baseline Model Details.}
In summary below, we use symbols to mark the model backbones. Here, $\dagger$ denotes models in ResNet50 \cite{He2015DeepRL} architecture and $\ast$ denotes models in WideResNet34 \cite{Zagoruyko16WRN}. The baseline models are obtained from top-ranked models of the same architecture in RobustBench \cite{croce2021robustbench}. 
\begin{itemize}[leftmargin=*,noitemsep,topsep=6pt]
    \item \textbf{Normal$^\dagger$/Normal$^\ast$}: Standard training.
    \item \textbf{Madry$_{\infty}^{\dagger}$}:  $\ell_{\infty}$ adversarial training~in \cite{madry2017towards}.
    \item \textbf{Trades$_{\infty}^{\ast}$}:  $\ell_{\infty}$ adversarial training~in \cite{zhang2019theoretically}. 
    \item \textbf{FAT$_{\infty}^{\ast}$}: \cite{zhang2020fat} uses friendly adversarial data that are confidently misclassified for adversarial training.
    
    \item \textbf{AWP$_{\infty}^{\ast}$}: \cite{wu2020adversarial} injects the worst-case weight perturbation during adversarial training to flatten the weight loss landscape.
    
    \item \textbf{PAT$_{self}^\dagger$}, \textbf{PAT$_{alex}^\dagger$}: Two adversarial training models based on the perceptual distance (LPIPS), two models differ: ResNet50 (\textit{self}) and AlexNet (\textit{alex}) \cite{laidlaw2021perceptual}.
    
    \item \textbf{Fast-AT$^\dagger$}: Computationally efficient $\ell_{\infty}$ adversarial training~in \cite{wong2020fast}.
\end{itemize}

\vspace{-4mm}
\paragraph{Training \& Evaluation Settings.}
We adopt the whole training set on both CIFAR-10 and ImageNet for model training.
In every training iterative step, the images in each batch share the same attack order. Besides, the Comp-PGD is applied on each image, where we set the iteration-update step $T$ as ten steps of each attack component for evaluation and seven steps for GAT. 
During the training of GAT, we apply every attack component on the input without the \textit{early-stopped} option to ensure the model could learn all attack components which have been launched. Furthermore, we evaluate two different order scheduling settings: \textit{random}/\textit{scheduled} during GAT on CIFAR-10.
Since both ordering mechanisms provide competitive robust models, therefore, we only use random ranking when training GAT on ImageNet, considering the training efficiency.
As mentioned in Sec. \ref{sub_sec:experiment_setup}, GAT utilizes a pre-trained model for fine-tuning to make the composite adversarial training more efficient than training from scratch. 
Different from the training phase of GAT, during the evaluation, we allow CAA to trigger the \textit{early-stop} option when the attack is successful, which can help us improve the attack success rate and reduce the computational cost.
Further discussion and comparison between different training settings of GAT, including using TRADES/Madry loss and fine-tuning/training from scratch, are given in Appendix \ref{sec:appendix_implementation_details}.

\vspace{-4mm}
\paragraph{Computing Resources and Code.}
For CIFAR-10, we train models on ResNet50 and WideResNet34 with SGD for 150 epochs. The training of GAT-f takes about 16 hours (ResNet50) and 28 hours (WideResNet34), and GAT-fs takes about 28 hours (ResNet50) and 55 hours (WideResNet34) on 8 Nvidia Tesla V100 GPUs. For ImageNet, we train ResNet50 with SGD for 100 epochs and about three days on 64 Nvidia Tesla V100 GPUs. The implementation is built with PyTorch \cite{Paszke19PyTorch}.

\vspace{-4mm}
\paragraph{Evaluation Metrics.} We report the models' Clean and Robust Accuracy (RA, \%) against multiple composite adversarial attacks. The RA aims to evaluate the model accuracy toward the fraction of perturbed examples retrieved from the test set which is correctly classified. We also provide the attack success rate (ASR, \%) in Appendix G, in which the higher indicates the stronger attack.

\input{preprint/assets/tables/cifar10_tables}
\input{preprint/assets/tables/imagenet_tables}

\subsection{Performance Evaluation}\label{subsec:exp}
\vspace{-3.5mm}The experimental results are shown in Table \ref{tab:cifar10_ra} (CIFAR-10) and Table \ref{tab:imagenet_ra} (ImageNet). On CIFAR-10, \textit{GAT-fs} and \textit{GAT-f} show competitive results. Both of them outperform all other baselines by a significant margin. For semantic attacks, the RA increases by 45\% $\sim$ 60\% on CIFAR-10, and 28\% $\sim$ 37\% on ImageNet. For full attacks, the RA increases by 15\% $\sim$ 27\% on CIFAR-10, and 9\% $\sim$ 15\% on ImageNet.
Nonetheless, the RA against three multiple threats with three different combinations, our proposed GAT keeps outperforming other baselines and shows the highest robustness of others. The comparison between GAT-f and GAT-fs demonstrates that GAT-fs can obtain higher RA against full attacks. However, the result also suggests a trade-off between the robustness of $\ell_{\infty}$ and semantic attacks.

Besides adversarial training models, we empirically observe that the RA of models with standard training has a degraded performance of 20\% $\sim$ 31\% on ImageNet data under semantic attacks (without $\ell_{\infty}$ attack). However, while $\ell_{\infty}$ attack is involved in the full attacks or other multiple threats (e.g., three attacks in Tables \ref{tab:cifar10_ra} and \ref{tab:imagenet_ra}), the models with only standard training are unable to resist these kinds of composite semantic perturbations, and the RA drops dramatically to 0\%.

\input{preprint/assets/figs/loss_landscape/five/five_imgs}

\subsection{Analysis, Discussion, and Visualization}\label{sub_sec:analysis}
\paragraph{Robust Accuracy vs. Number of Attacks and Their combinations.}
We conduct an ablation study to show that the number of attacks and their combinations can hugely affect robust accuracy, illustrating the importance of attack ordering and the new insights into robustness through our proposed composite adversarial examples.
Fig. \ref{fig:composite_attack_eval} already demonstrates that our model is the most resilient to composite adversarial examples consisting of different numbers of attacks, in terms of attaining the lowest attack success rate in the test set that each model initially correctly classified. Furthermore, Table \ref{tab:ra_with_diff_attack_num} shows that as the number of attacks increases (\textit{CAA}$_{1}$ to \textit{CAA}$_{6}$), the RA of our proposed GAT consistently outperforms all other models.
Specifically, GAT outperforms other baselines by up to 35\%. Although the standard model (Normal$^\dagger$) has the advantage of higher cleaning accuracy, it is still not resistant to semantic and various composite adversarial perturbations.
Results of \textit{three attacks} in Tables \ref{tab:cifar10_ra} and \ref{tab:imagenet_ra} demonstrate the effect of different combinations when the number of attacks is fixed. Comparing GAT with others on both CIFAR-10 and ImageNet, the result shows that \textit{\textit{GAT-f}} is more robust than all baselines under three different attacks by 9\% $\sim$ 23\%. On ImageNet, \textit{GAT-f} also outperforms those baselines. For more experimental results, including single attacks, Auto-attack, two-component attacks, and other results on other datasets (e.g., SVHN),  please refer to Appendix \ref{sec:appendix_additional_exp}.

\input{preprint/assets/tables/figure1b_ra}

\vspace{-4mm}\paragraph{Effectiveness of Random vs. Scheduled Ordering.} We evaluated the effectiveness of random versus scheduled ordering by conducting pairwise t-tests. Ten experiments were conducted with different initializations, and the experimental results on CIFAR10/Full-attack demonstrated that the robust accuracy of the \textit{scheduled} ordering was significantly lower than that of the \textit{random} ordering ($p$-value $<.001$ for all models).

\vspace{-4mm}\paragraph{Inadequacies of Current Adversarial Robustness Assessments.}
Existing methods for evaluating adversarial robustness, which only considers perturbations in $\ell_{p}$-ball, may be incomplete and biased. To investigate this issue, we compared the rankings of the top ten models on the RobustBench dataset (CIFAR-10, $\ell_{\infty}$) \cite{croce2021robustbench}. We found that the rankings between Auto-Attack and CAA had a low correlation, suggesting the need for more comprehensive assessments. Specifically, we computed the Spearman’s rank correlation coefficient between Auto-Attack and CAA (rand. \& sched.) for semantic and full attacks, which yielded values of 0.16 (rand. \emph{vs.} sched.) and 0.36 (rand. \emph{vs.} Auto) and 0.38 (sched. \emph{vs.} Auto), respectively. These findings underscore the importance of developing new methods that can more accurately evaluate adversarial robustness.

\vspace{-4mm}\paragraph{Visualization of Loss Landscape.}
To gain a deeper understanding of why our proposed approach leads to significant improvements in adversarial robustness, we visualized the loss landscape of a single semantic attack under three different models: standardly trained ResNet50 (Normal$^{\dagger}$), ResNet50 with $\ell_{\infty}$-robust training (Madry$_{\infty}^{\dagger}$), and our proposed GAT approach (\textit{GAT-f}$^{\dagger}$), see Fig. \ref{fig:loss_singles}. We plotted the cross-entropy loss of selected samples for each model, sweeping over the semantic perturbation space within a designated interval. We empirically observe that across five different single semantic attacks, the curves (green) generated by GAT were much smoother, flatter, and lower than those produced by the other models. We believe that this phenomenon sheds light on the effectiveness of our proposed approach, which can indeed train a model robust to the composite adversarial perturbations.

%% file: preprint/assets/tables/cifar10_tables.tex
\begin{table*}[!htb]
\setlength\tabcolsep{4.8pt}
\centering
    \begin{tabular}{p{4.1em}|r|rrr|rr|rr}
    \toprule
     &   & \multicolumn{3}{c|}{Three attacks} & \multicolumn{2}{c|}{Semantic attacks}  & \multicolumn{2}{c}{Full attacks}\\
    Training  & Clean & \textit{CAA}$_{3a}$ & \textit{CAA}$_{3b}$ & \textit{CAA}$_{3c}$ & Rand. & Sched. & Rand. & Sched. \\
    \midrule
    Normal$^\dagger$ & 95.2  & 0.0 {\footnotesize $\pm$ 0.0} & 0.0 {\footnotesize $\pm$ 0.0} & 0.0 {\footnotesize $\pm$ 0.0} & 59.7 {\footnotesize $\pm$ 0.2} & 44.2 {\footnotesize $\pm$ 0.5} & 0.0 {\footnotesize $\pm$ 0.0} & 0.0 {\footnotesize $\pm$ 0.0} \\
    Madry$_{\infty}^{\dagger}$ & 87.0  & 30.8 {\footnotesize $\pm$ 0.2} & 18.8 {\footnotesize $\pm$ 0.5} & 19.1 {\footnotesize $\pm$ 0.3} & 31.5 {\footnotesize $\pm$ 0.2} & 21.3 {\footnotesize $\pm$ 0.2} & 10.8 {\footnotesize $\pm$ 0.2} & 3.7 {\footnotesize $\pm$ 0.2} \\
    PAT$_{self}^\dagger$ & 82.4  & 20.9 {\footnotesize $\pm$ 0.1} & 11.9 {\footnotesize $\pm$ 0.5} & 17.9 {\footnotesize $\pm$ 0.3} & 28.9 {\footnotesize $\pm$ 0.3} & 17.5 {\footnotesize $\pm$ 0.3} & 9.1 {\footnotesize $\pm$ 0.3} & 2.5 {\footnotesize $\pm$ 0.3} \\
    PAT$_{alex}^\dagger$ & 71.6  & 20.7 {\footnotesize $\pm$ 0.3} & 12.5 {\footnotesize $\pm$ 0.2} & 16.5 {\footnotesize $\pm$ 0.4} & 23.4 {\footnotesize $\pm$ 0.3} & 12.2 {\footnotesize $\pm$ 0.4} & 10.3 {\footnotesize $\pm$ 0.1} & 2.5 {\footnotesize $\pm$ 0.2} \\
    \textbf{GAT-f}$^\dagger$ & \textbf{82.3} & \textbf{39.9 {\footnotesize $\pmb{\pm}$ 0.1}} & \textbf{33.3 {\footnotesize $\pmb{\pm}$ 0.1}} & \textbf{28.9 {\footnotesize $\pmb{\pm}$ 0.2}} & \textbf{69.9 {\footnotesize $\pmb{\pm}$ 0.1}} & \textbf{66.0 {\footnotesize $\pmb{\pm}$ 0.1}} & \textbf{30.0 {\footnotesize $\pmb{\pm}$ 0.4}} & \textbf{18.8 {\footnotesize $\pmb{\pm}$ 0.3}} \\
    \textbf{GAT-fs}$^\dagger$ & \textbf{82.1} & \textbf{43.5 {\footnotesize $\pmb{\pm}$ 0.1}} & \textbf{36.6 {\footnotesize $\pmb{\pm}$ 0.1}} & \textbf{32.5 {\footnotesize $\pmb{\pm}$ 0.1}} & \textbf{69.9 {\footnotesize $\pmb{\pm}$ 0.2}} & \textbf{66.6 {\footnotesize $\pmb{\pm}$ 0.1}} & \textbf{32.3 {\footnotesize $\pmb{\pm}$ 0.8}} & \textbf{21.8 {\footnotesize $\pmb{\pm}$ 0.3}} \\
    \midrule
    Normal$^\ast$ & 94.0  & 0.0 {\footnotesize $\pm$ 0.0} & 0.0 {\footnotesize $\pm$ 0.0} & 0.0 {\footnotesize $\pm$ 0.0} & 46.0 {\footnotesize $\pm$ 0.4} & 29.9 {\footnotesize $\pm$ 0.5} & 0.0 {\footnotesize $\pm$ 0.0} & 0.0 {\footnotesize $\pm$ 0.0} \\
    Trades$_{\infty}^\ast$ & 84.9  & 30.0 {\footnotesize $\pm$ 0.3} & 19.8 {\footnotesize $\pm$ 0.6} & 10.1 {\footnotesize $\pm$ 0.5} & 16.6 {\footnotesize $\pm$ 0.2} & 8.1 {\footnotesize $\pm$ 0.5} & 5.8 {\footnotesize $\pm$ 0.3} & 1.5 {\footnotesize $\pm$ 0.2} \\
    FAT$_{\infty}^\ast$ & 88.1  & 29.8 {\footnotesize $\pm$ 0.4} & 17.1 {\footnotesize $\pm$ 0.4} & 12.8 {\footnotesize $\pm$ 0.6} & 18.7 {\footnotesize $\pm$ 0.2} & 9.8 {\footnotesize $\pm$ 0.5} & 6.1 {\footnotesize $\pm$ 0.1} & 1.5 {\footnotesize $\pm$ 0.1} \\
    AWP$_{\infty}^\ast$ & 85.4  & 34.2 {\footnotesize $\pm$ 0.2} & 23.2 {\footnotesize $\pm$ 0.2} & 11.1 {\footnotesize $\pm$ 0.4} & 15.6 {\footnotesize $\pm$ 0.2} & 7.9 {\footnotesize $\pm$ 0.2} & 5.9 {\footnotesize $\pm$ 0.0} & 1.7 {\footnotesize $\pm$ 0.2} \\
    \textbf{GAT-f}$^\ast$ & \textbf{83.4} & \textbf{40.2 {\footnotesize $\pmb{\pm}$ 0.1}} & \textbf{34.0 {\footnotesize $\pmb{\pm}$ 0.1}} & \textbf{30.7 {\footnotesize $\pmb{\pm}$ 0.4}} & \textbf{71.6 {\footnotesize $\pmb{\pm}$ 0.1}} & \textbf{67.8 {\footnotesize $\pmb{\pm}$ 0.2}} & \textbf{31.2 {\footnotesize $\pmb{\pm}$ 0.4}} & \textbf{20.1 {\footnotesize $\pmb{\pm}$ 0.3}} \\
    \textbf{GAT-fs}$^\ast$ & \textbf{83.2} & \textbf{43.5 {\footnotesize $\pmb{\pm}$ 0.1}} & \textbf{36.3 {\footnotesize $\pmb{\pm}$ 0.1}} & \textbf{32.9 {\footnotesize $\pmb{\pm}$ 0.4}} & \textbf{70.5 {\footnotesize $\pmb{\pm}$ 0.1}} & \textbf{66.7 {\footnotesize $\pmb{\pm}$ 0.3}} & \textbf{32.2 {\footnotesize $\pmb{\pm}$ 0.7}} & \textbf{21.9 {\footnotesize $\pmb{\pm}$ 0.7}} \\
    \bottomrule
    \end{tabular}
    \vspace{-1mm}
    \caption{Comparison of accuracy (\%) on CIFAR-10. We combine different types of three attacks (\textit{CAA}$_{3}$) with scheduled ordering: \textit{CAA}$_{3a}$: (Hue, Saturation, $\ell_{\infty}$), \textit{CAA}$_{3b}$: (Hue, Rotation, $\ell_{\infty}$), \textit{CAA}$_{3c}$: (Brightness, Contrast, $\ell_{\infty}$), on CIFAR-10}
\vspace{-2mm}
    \label{tab:cifar10_ra}%
\end{table*}

%% file: preprint/assets/tables/imagenet_tables.tex
\begin{table*}[!htb]
\setlength\tabcolsep{4.8pt}
\centering
    \begin{tabular}{p{4.1em}|r|rrr|rr|rr}
    \toprule
     &  & \multicolumn{3}{c|}{Three attacks} & \multicolumn{2}{c|}{Semantic attacks}  & \multicolumn{2}{c}{Full attacks}\\
    Training  & Clean & \textit{CAA}$_{3a}$ & \textit{CAA}$_{3b}$ & \textit{CAA}$_{3c}$ & Rand. & Sched. & Rand. & Sched. \\
    \midrule
Normal$^\dagger$ & 76.1  & 0.0 {\footnotesize $\pm$ 0.0} & 0.0 {\footnotesize $\pm$ 0.0} & 0.0 {\footnotesize $\pm$ 0.0} & 31.2 {\footnotesize $\pm$ 0.4} & 20.6 {\footnotesize $\pm$ 1.0} & 0.0 {\footnotesize $\pm$ 0.0} & 0.0 {\footnotesize $\pm$ 0.0} \\
Madry$_{\infty}^{\dagger}$ & 62.4  & 13.9 {\footnotesize $\pm$ 0.4} & 9.2 {\footnotesize $\pm$ 0.2} & 16.2 {\footnotesize $\pm$ 0.8} & 14.0 {\footnotesize $\pm$ 0.1} & 9.0 {\footnotesize $\pm$ 0.0} & 7.1 {\footnotesize $\pm$ 0.1} & 2.8 {\footnotesize $\pm$ 0.2} \\
Fast-AT$_{\infty}^\dagger$ & 53.8  & 9.5 {\footnotesize $\pm$ 0.3} & 5.5 {\footnotesize $\pm$ 0.1} & 11.4 {\footnotesize $\pm$ 0.8} & 6.3 {\footnotesize $\pm$ 0.1} & 3.6 {\footnotesize $\pm$ 0.1} & 3.1 {\footnotesize $\pm$ 0.1} & 1.0 {\footnotesize $\pm$ 0.1} \\
\textbf{GAT-f}$^\dagger$ & 60.0  & \textbf{19.2 {\footnotesize $\pmb{\pm}$ 1.0}} & \textbf{18.9 {\footnotesize $\pmb{\pm}$ 1.4}} & \textbf{18.4 {\footnotesize $\pmb{\pm}$ 0.4}} & \textbf{43.5 {\footnotesize $\pmb{\pm}$ 1.9}} & \textbf{38.9 {\footnotesize $\pmb{\pm}$ 2.0}} & \textbf{18.5 {\footnotesize $\pmb{\pm}$ 0.5}} & \textbf{11.8 {\footnotesize $\pmb{\pm}$ 0.1}} \\
    \bottomrule
    \end{tabular}
    \vspace{-1mm}
    \caption{Comparison of accuracy (\%) on ImageNet. (CAA$_{3a,3b,3c}$: same combination as Table \ref{tab:cifar10_ra})}
\vspace{-3mm}
    \label{tab:imagenet_ra}%
\end{table*}

%% file: preprint/assets/figs/loss_landscape/five/five_imgs.tex
\begin{figure*}[htbp]
  \subfloat[Hue]{
	\begin{minipage}[c]{.192\linewidth}
	   \centering
	   \includegraphics[width=\linewidth, trim=2cm 0.05cm 2cm 0.05cm, clip]{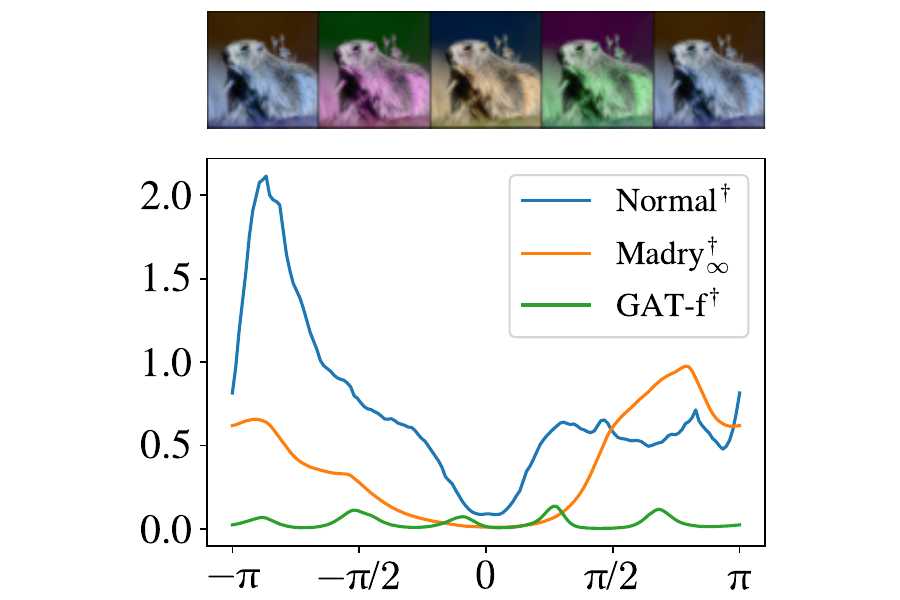}
	\end{minipage}}
 \hfill
  \subfloat[Rotation]{
	\begin{minipage}[c]{.192\linewidth}
	   \centering
	   \includegraphics[width=\linewidth, trim=2cm 0.05cm 2cm 0.05cm, clip]{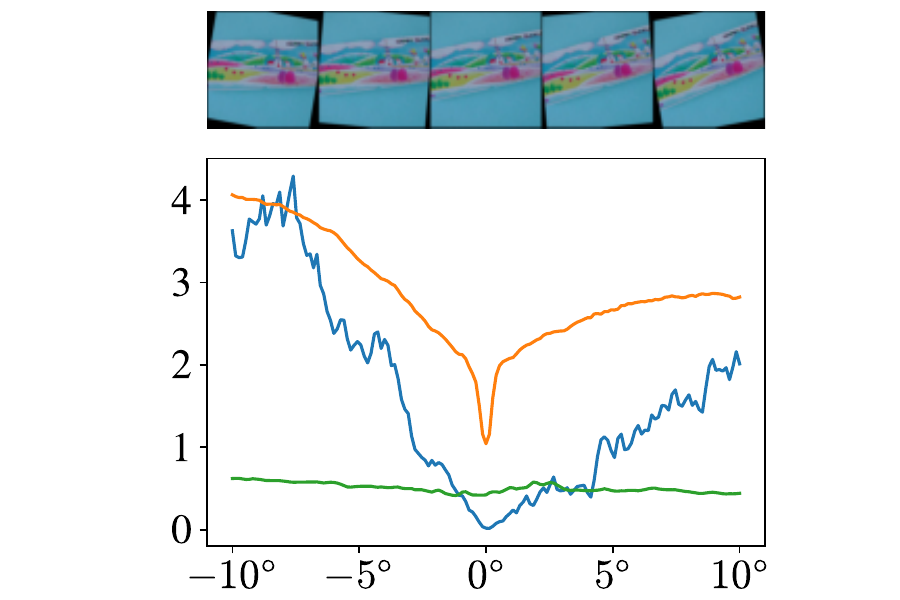}
	\end{minipage}}
 \hfill
  \subfloat[Saturation]{
	\begin{minipage}[c]{.192\linewidth}
	   \centering
	   \includegraphics[width=\linewidth, trim=2cm 0.05cm 2cm 0.05cm, clip]{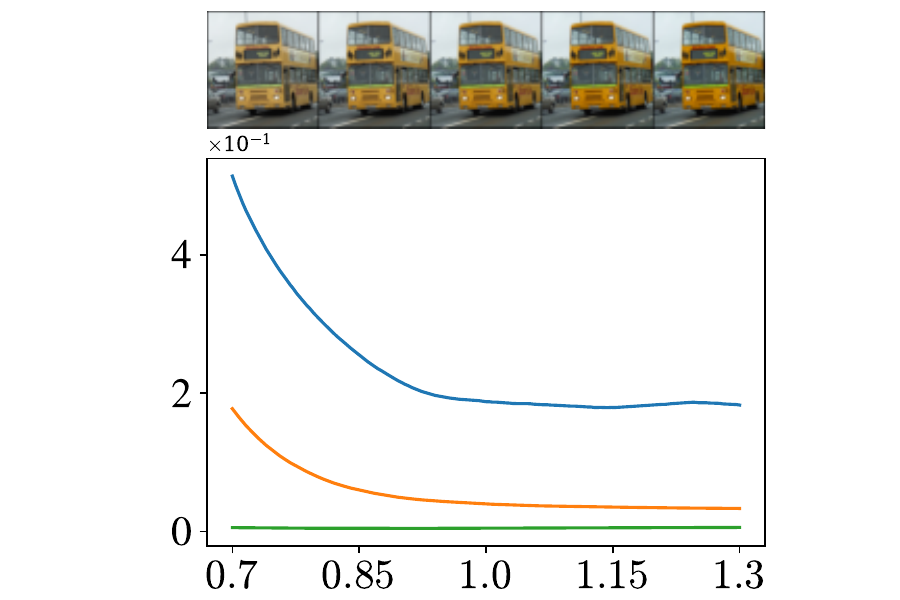}
	\end{minipage}}
 \hfill
  \subfloat[Brightness]{
	\begin{minipage}[c]{.192\linewidth}
	   \centering
	   \includegraphics[width=\linewidth, trim=2cm 0.05cm 2cm 0.05cm, clip]{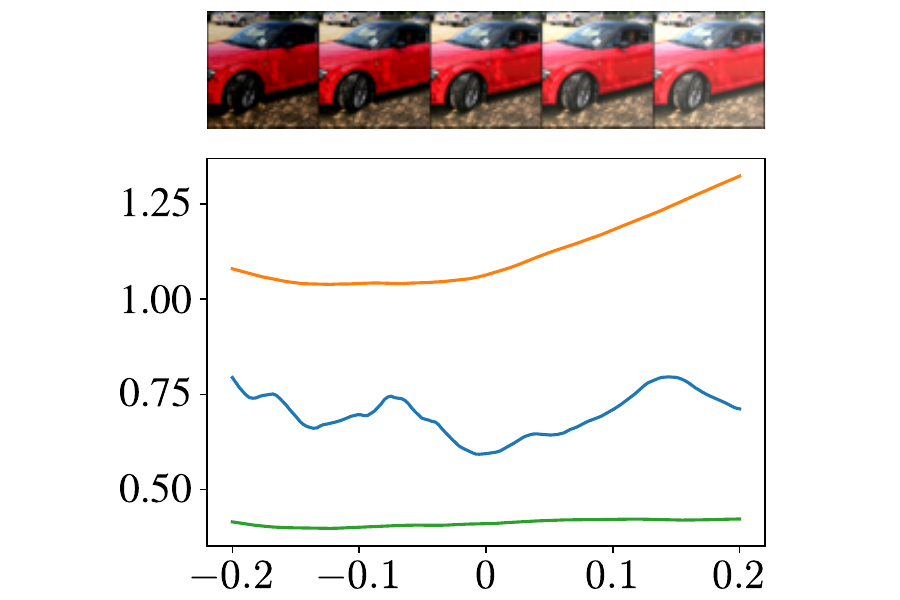}
	\end{minipage}}
 \hfill
  \subfloat[Contrast]{
	\begin{minipage}[c]{.192\linewidth}
	   \centering
	   \includegraphics[width=\linewidth, trim=2cm 0.05cm 2cm 0.05cm, clip]{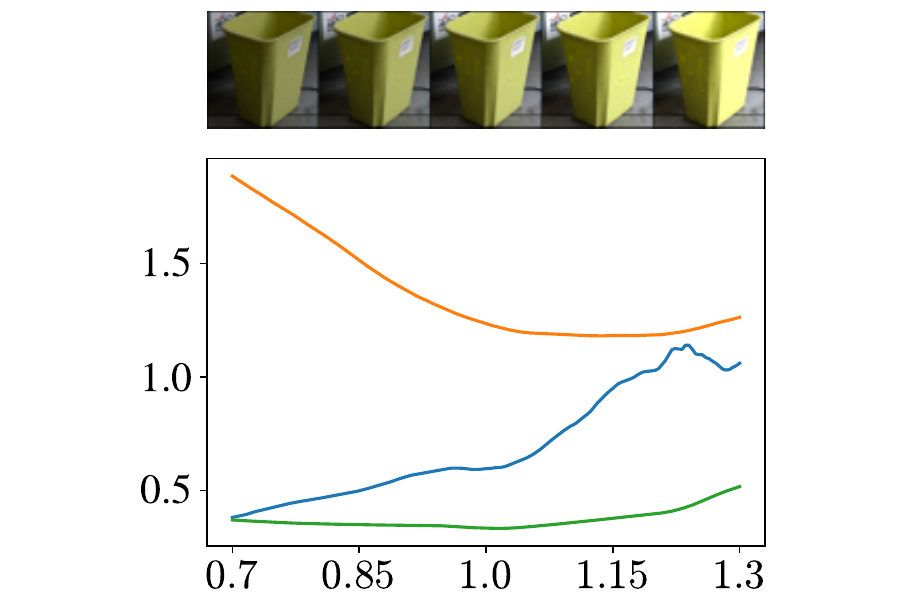}
	\end{minipage}}
\vspace{-1mm}
  \caption{Loss landscape of selected examples when performing five different semantic attacks under models produced by different adversarial training approaches. The blue and orange color curves represent standard and $\ell_{\infty}$ robust model respectively; the green color curve represents GAT-f model.}
  \label{fig:loss_singles}
    \vspace{-3mm}
\end{figure*}

%% file: preprint/assets/tables/figure1b_ra.tex
\begin{table}[t]
\setlength\tabcolsep{4.15pt}
\centering
    \begin{tabular}[table-number-alignment = center]{l|rrrrrr}
    \toprule
    Training & \textit{CAA}$_1$ & \textit{CAA}$_2$ & \textit{CAA}$_3$ & \textit{CAA}$_4$ & \textit{CAA}$_5$ & \textit{CAA}$_6$ \\
    \midrule
Normal$^\dagger$ & 50.9  & 45.8  & 33.4  & 25.8  & 21.1  & 0.0  \\
Madry$_{\infty}^{\dagger}$ & 38.1  & 33.9  & 21.9  & 14.4  & 9.0   & 2.8  \\
Fast-AT$_{\infty}^\dagger$ & 27.8  & 23.9  & 12.7  & 7.0   & 3.6   & 1.0  \\
\textbf{GAT-f}$^\dagger$ & \textbf{51.0} & \textbf{48.2} & \textbf{44.5} & \textbf{42.2} & \textbf{38.9} & \textbf{11.8} \\
\bottomrule
\end{tabular}%
    \caption{Comparison of RA (\%) on four adversarial training approaches against six different \textit{CAAs} on ImageNet. \textit{CAA}$_{1}$: (Hue), \textit{CAA}$_{2}$: (Hue, Saturation), \textit{CAA}$_{3}$: (Hue, Saturation, Rotation), \textit{CAA}$_{4}$: (Hue, Saturation, Rotation, Brightness), \textit{CAA}$_{5}$: (Hue, Saturation, Rotation, Brightness, Contrast), \textit{CAA}$_{6}$: (Hue, Saturation, Rotation, Brightness, Contrast, $\ell_\infty$)}
    \label{tab:ra_with_diff_attack_num}%
    \vspace{-4mm}
\end{table}

%% file: preprint/docs/5_conclusion.tex
\section{Conclusion}\label{sec:conslusions}
In this paper, we proposed GAT, a generic approach for enhancing the robustness of deep learning models to composite semantic perturbations, with the ultimate goal of preparing classifiers for the real world. Our approach is based on a unique design of attack order scheduling for multiple perturbation types and the optimization of each attack component. This further enables GAT to achieve state-of-the-art robustness against a wide range of adversarial attacks, including those in $\ell_{p}$ norms and semantic spaces. Evaluated on CIFAR-10 and ImageNet datasets, our results demonstrate that GAT achieves the highest robust accuracy on most composite attacks by a large margin, providing new insights into achieving compositional adversarial robustness. We believe our work sheds new light on the frontiers of realistic adversarial attacks and defenses.

%% file: preprint/docs/6_appendix.tex
\section*{Appendix}\label{sec:appendix}
\section{Implementation Details}\label{sec:appendix_implementation_details}

\paragraph{Training Phase.}
\input{preprint/assets/tables/epsilons}
In the implementation of generalized adversarial training (GAT), we consider two model architectures, ResNet-50 \cite{He2015DeepRL} and WideResNet-34 \cite{Zagoruyko16WRN}, on CIFAR-10 dataset \cite{Krizhevsky09learningmultiple}; and ResNet-50 on ImageNet dataset \cite{ILSVRC15}. For CIFAR-10, we set the maximum training epoch to 150 with the batch size 2048 and selected the model with the best evaluation test accuracy. The learning rate is set to 0.1 at the beginning and exponentially decays. We utilize the warming-up learning rate technique for the first ten epochs, which means the learning rate would linearly increase from zero to the preset value (0.1) in the first ten epochs.
For ImageNet, we set the maximum training epoch to 100 with the batch size 1536 and selected the model with the best evaluation test accuracy. The learning rate is set to 0.1 at the beginning and exponentially decays by 0.1 every 30 epochs. Similarly, we utilize the warming-up learning rate technique for the first five epochs. 
We launched all threat models (full attacks) while training; for each batch, we utilized scheduled ordering for \textit{GAT-fs} and random ordering for \textit{GAT-f}. The iteration step $T$ of each attack for Comp-PGD is set to 7, and the step size of attack $A_k$ is set as $2.5 \cdot (\beta_k - \alpha_k) / 2T$, where $\beta_k$ and $\alpha_k$ are the values of perturbation intervals defined in Table \ref{tab:attack_epsilons}. 

\paragraph{Testing Phase.}
To compare our GAT approach with other adversarial training baselines, we launch composite adversarial attacks (CAAs) of different numbers of attack types, including single attacks, two attacks, three attacks, all semantic attacks, and full attacks on each robust model. Furthermore, the iteration step $T$ of each attack for Comp-PGD is set as 10, and the step size is the same as the training settings. In addition, the maximum iteration of \textit{order scheduling} is designated as five, and we will launch the early-stop option in every update step while the CAA succeeds in attacking. Note that the ASR would slightly decrease ($\approx${\footnotesize~}2\%) if the early-stop feature is disabled. This is likely due to the highly complex and non-convex loss landscape (Fig. \ref{fig:appendix_loss_traces}); while the early-stop feature helps CAA maintain its attack efficiency.

\paragraph{Training Strategy.}
Our training process considers two training strategies: 1) training from scratch and 2) fine-tuning on $\ell_\infty$-robust models; two learning objectives: 1) Madry's loss \cite{madry2017towards} and 2) TRADES' loss \cite{zhang2019theoretically}. Note that $x_{\text{c-adv}}\in{\mathcal{B}(x;\Omega;E)}$ denotes the composite adversarial example $x_\text{c-adv}$ is perturbed by attacks from $\Omega$ within the perturbation intervals $E$. The main difference between these two is shown in Eq. \ref{eqn:madry_loss} and Eq. \ref{eqn:trades_loss}. That is, Eq. \ref{eqn:trades_loss} encourages the natural error to be optimized in the first term; meanwhile, the robust error in the second regularization term could help minimize the distance between the prediction of natural samples and adversarial samples. Zhang et al. theoretically proved that this design of loss function could help the outputs of the model to be smooth \cite{zhang2019theoretically}.

\begin{align}
    \label{eqn:madry_loss}
    \min_{\theta_{\mathcal{F}}}\mathop{\mathbb{E}}_{(x,y)\sim\mathcal{D}}
    \bigg[
    \max_{x_{\text{c-adv}}\in{\mathcal{B}(x;\Omega;E)}} \mathcal{L}_{ce}(\mathcal{F}(x_{\text{c-adv}}),y)\bigg]
\end{align}

\begin{align}
    \label{eqn:trades_loss}
    \min_{\theta_{\mathcal{F}}}\mathop{\mathbb{E}}_{(x,y)\sim\mathcal{D}}\bigg[\mathcal{L}(\mathcal{F}(x), y) + \beta\cdot\max_{x_{\text{c-adv}}\in{\mathcal{B}(x;\Omega;E)}} \mathcal{L}(\mathcal{F}(x),\mathcal{F}(x_{\text{c-adv}}))\bigg]
\end{align}

\begin{figure}[ht]
  \begin{subfigure}[b]{.48\textwidth}
      \centering
      \includegraphics[width=\linewidth, trim=0.5cm 0.2cm 1.8cm 1.2cm, clip]{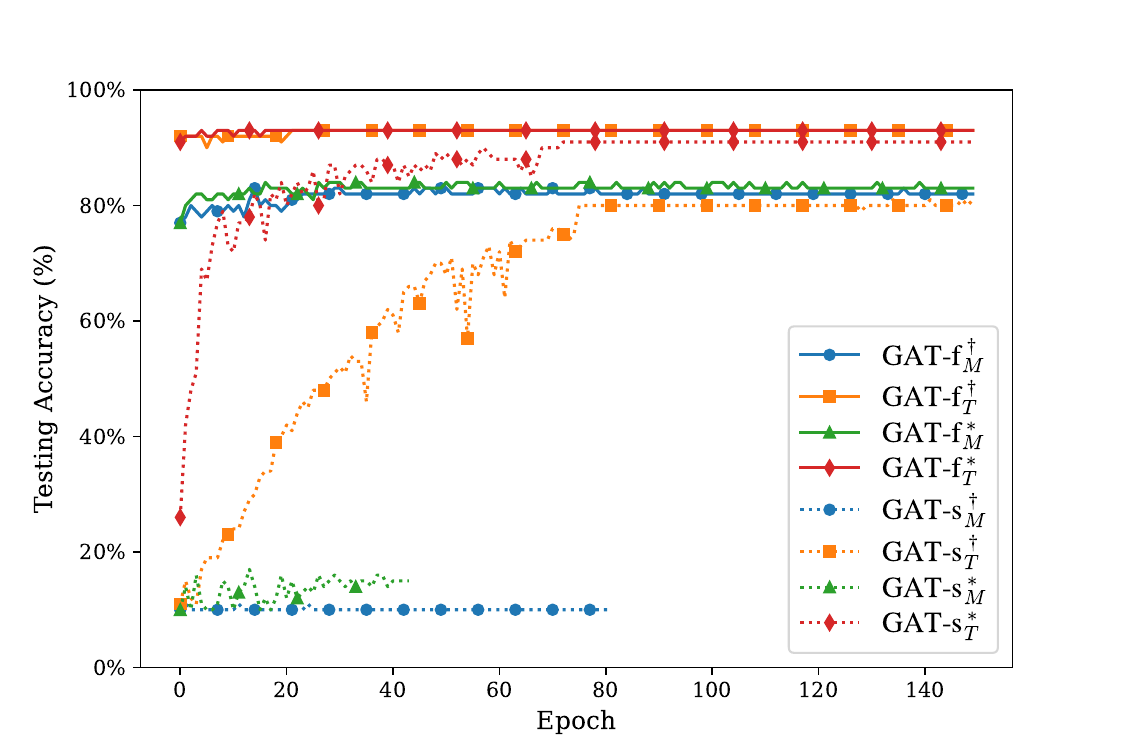}
      \caption{}
      \label{fig:cifar10_training_records}
  \end{subfigure}
  \hfill
  \begin{subfigure}[b]{.48\textwidth}
      \centering
      \includegraphics[width=\linewidth, trim=1.2cm 0cm 1.2cm 0.1cm, clip]{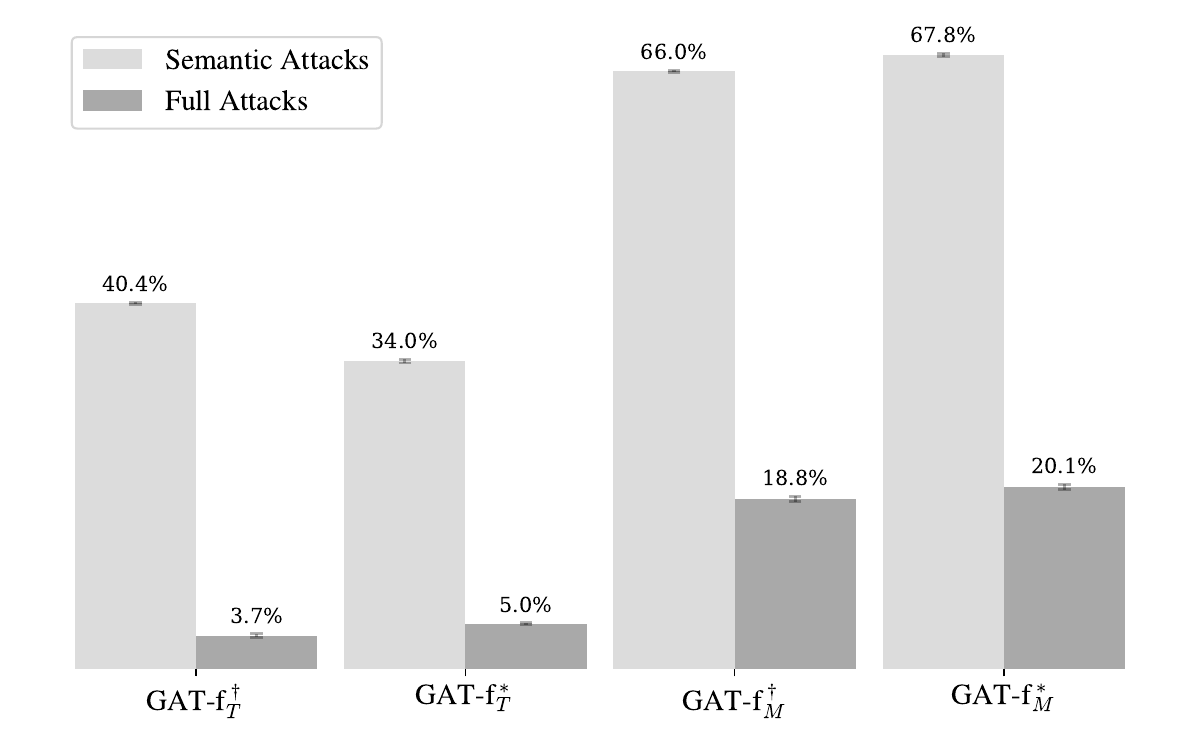}
      \caption{}
      \label{fig:cifar_finetune_models_ra}
  \end{subfigure}\vspace{-2mm}
  \caption{(a) The testing accuracy during generalized adversarial training on CIFAR-10. The models differ in different training scenarios; the lower script $T$ denotes that the model using \textit{TRADES}' loss \cite{zhang2019theoretically} for training, and $M$ for \textit{Madry}'s loss \cite{madry2017towards}. The upper script $\dagger$ denotes the model using ResNet50 \cite{He2015DeepRL} backbone and $\ast$ is for WideResNet34 \cite{Zagoruyko16WRN}. (b) The robust accuracy (\%) of our GAT fine-tuned models under semantic and full attacks.
  }\vspace{-2mm}
\end{figure}

As shown in Fig. \ref{fig:cifar10_training_records}, we evaluate the clean test accuracy of GAT models in every epoch with different training settings, including using two architectures (ResNet-50 / WideResNet-34), two learning objectives, and two training strategies mentioned above. We empirically find the models using fine-tuning strategy (solid curves) can achieve higher clean test accuracy than most of the models training from scratch (dotted curves). Furthermore, we evaluate the robust test accuracy for these four models (see Fig. \ref{fig:cifar_finetune_models_ra}). Under the semantic and full attacks, the models GAT-f$_M$ (fine-tuning with Madry's loss) achieve higher robust accuracy than GAT-f$_T$ (fine-tuning with TRADES loss). Hence, in the section of experimental results, we utilized the GAT models, which are trained with Madry's loss and fine-tuning on a $\ell_\infty$-robust model.
\clearpage

\vspace{-2mm}\section{Algorithm of the Composite Adversarial Attack (CAA)}\label{sec:appendix_gat_algo}
\input{preprint/assets/gat_algorithm}
\vspace{-4mm}
\clearpage

\section{The Loss Trace Analysis of Component-wise PGD (Comp-PGD)}\label{sec:appendix_loss_landscape}
To demonstrate the effectiveness of Comp-PGD, in Fig. \ref{fig:appendix_loss_traces}, we visualize the update process of Comp-PGD when performing a single semantic attack on the WideResnet-34 model.
We uniformly sample 20 start points for each attack and update $\delta_k$ using Comp-PGD by these initial points. The red margins of each sub-figure in Fig. \ref{fig:appendix_loss_traces} represent the margin of successful attack by our samples. The endpoints of the loss trace show obviously that Comp-PGD indeed can help search for the worst case by maximizing the loss during each attack.
\input{preprint/assets/figs/loss_landscape/loss_traces}

\section{Ablation Study: Attack Components' Optimization}\label{sec:appendix_sep_opt}
\subsection{Why Separately Optimize the Attack Parameters? (Comp-PGD vs. Ensemble-PGD)}
In this paper, we used Comp-PGD to optimize the individual attack component. On the other hand, one can also optimize all attack components simultaneously given an attack order, for which we call \textit{Ensemble-PGD}. Specifically, CAA can jointly optimize the attack parameters for an attack chain \textit{at a chosen fixed attack order}. In this regard, we repeated the same experiments on CIFAR-10 but considered optimizing the attack parameters \textit{simultaneously} instead of \textit{sequentially}. The results show that Ensemble-PGD does not provide better attack capacity (see Table \ref{tab:cifar10_ablation}) than Comp-PGD (see Table \ref{tab:appendix_cifar10_multi_asr}). We provide the experimental results in Attack Success Rate (ASR), as it represents the strength of the attack (higher means a more vigorous attack). Although GAT approaches still outperform other baselines in defending against all threats, the results showed that Ensemble-PGD generally has \textit{lower} attack performance than Comp-PGD. This is probably due to the fact that the number of the variables for optimizing in Ensemble-PGD is higher than that of Comp-PGD (in each sequential step), making the optimization process harder to achieve similar results.
\input{preprint/assets/tables/ablation1}

\subsection{Why Not Optimize Attack Power by Grid Search? (Comp-PGD vs. Grid Search)}
It is intuitive to optimize the attack parameters (levels) in a brute-force way, i.e., \textit{grid search}. However, doing so would exponentially increase the computational cost as the number of attacks increases. We conduct an experiment to compare the attack success rate (ASR, \%) between the Grid-Search attack and our proposed CAA. We include all types of semantic attacks (Hue, Saturation, Rotation, Brightness, and Contrast) in this experiment. Also, since there are $N!$ kinds of attack orders for $N$ attacks, for simplicity, we chose only one attack order here and utilized the same attack order in CAA (fixed).

\begin{figure}[h]
    \centering
    \includegraphics[width=\linewidth, trim=0.5cm 0.25cm 0.43cm 0.2cm, clip]{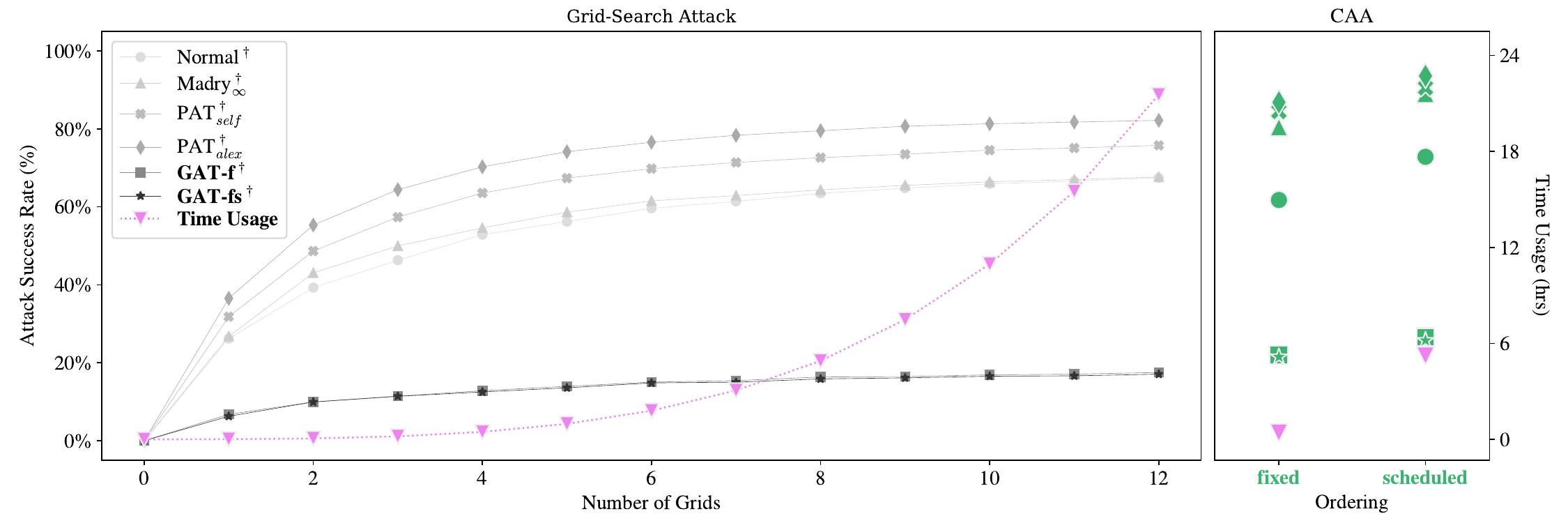}
    \vspace{-2mm}
    \caption{Comparison of attack success rate between Grid-Search Attack and CAA.}
    \label{fig:grid_search_vs_caa}
\end{figure}

As shown in Fig. \ref{fig:grid_search_vs_caa}, the results demonstrated that CAA is obviously stronger than grid search, with a significantly lower computational cost. The results also indicate that CAA is more valuable than grid-search-based optimization, as CAA consistently achieves a higher ASR. This is because, in grid search, it could only look into the discrete attack value space; clearly, it would need to increase spatial density (grid numbers) to obtain a higher attack success rate. To be more specific, given the grid numbers $K$ (uniformly sampled points in each attack space), the attack complexity of Grid-Search Attack is $\mathcal{O}(K^N)$; and the attack complexity of CAA (fixed order) is $\mathcal{O}(N\cdot T\cdot R)$, where $T$ is the optimization steps for Comp-PGD, and $R$ is the number of restarts. That is, we allow CAA to optimize each attack with $R$ different starting points. In our experiment, since CAA could search for the optimal attack value by gradient-based search, we need only five restart points ($R$) and ten steps for Comp-PGD optimization ($T$) to outperform the grid-search-based strategy. In this scenario, the attack complexity of Grid-Search Attack is higher than CAA  (since $\mathcal{O}(K^N)>\mathcal{O}(N\cdot K^2)>\mathcal{O}(N\cdot T\cdot R)$, given $T, R\leq K$).

\vspace{-4mm}
\section{Ablation Study: Order Scheduling and Comp-PGD Are Essential to Strengthen GAT}\label{sec:appendix_random_training}
\vspace{-2mm}
To further verify that our scheduling mechanism and Comp-PGD play essential roles in CAA while doing GAT, we remove the order scheduling feature and Comp-PGD but pre-generate training data by adding random semantic perturbations on the CIFAR-10 training set, referring to RSP-10. That is, RSP-10 is generated in random attack ordering and random attack parameters on CIFAR-10. We then performed regular adversarial training on RSP-10 to obtain the robust models \cite{zhang2019theoretically}, including from-scratch and fine-tuning. Table \ref{tab:appendix_rsp_cifar10_ra} listed the robust accuracy of three such robust models under three attacks, semantic attacks and full attacks. The results show that GAT still outperforms other baselines for up to 27\%/54\%/25\% in three/semantic/full attacks, demonstrating that order scheduling and Comp-PGD are essential to harden GAT to derive a robust model.
\input{preprint/assets/tables/ablation2}

\clearpage

\section{Sensitive Analysis and Additional Discussions}\label{sec:appendix_sen_analysis_order}
\subsection{The Attack Order Matters! The Two-attack Experiments}\label{subsec:two_attack}
We conduct an analysis on different \textit{order} types under two attacks to demonstrate the influence of order on CAA. As shown in Table \ref{tab:appendix_imagenet_two_asr}, we list the attack success rate (ASR) of two attacks with different orders ($\ell_\infty\to$ \textit{semantic} attack / \textit{semantic} attack $\to\ell_\infty$) on GAT and other baseline models. The results show that most baselines are more fragile to the CAA with a semantic attack launched first than the attack with $\ell_\infty$ first. Furthermore, \textit{GAT-f} has the smallest ASR change when alternating the order, indicating that GAT helps improve the robustness when the attack order is changed. 

\input{preprint/assets/tables/imagenet/two_asr}

\vspace{-2mm}
\definecolor{correct}{rgb}{0, 0.615, 0}
\definecolor{incorrect}{rgb}{1, 0.333, 0.333}
\subsection{How Do Composite Perturbations Fool the Model? Visual Examples}
In Fig. \ref{fig:appendix_order_effect_1}, we present the inference results from an $\ell_\infty$-robust model (Madry$_{\infty}^{\dagger}$); the confidence bars are marked in \textcolor{correct}{green} (\textcolor{incorrect}{red}) if the prediction is correct (incorrect).
The results showed that while a robust model can resist perturbations in $\ell_{p}$ ball, this only consideration is not comprehensive. That is, if we consider computing $\ell_{\infty}$ perturbations after some semantic attacks, the model may not exhibit the robustness it has around the $\ell_{\infty}$ ball.

\vspace{-1mm}
\input{preprint/assets/figs/order-effect/order-effect}

\clearpage
\section{Additional Experimental Results and Adversarial Examples}\label{sec:appendix_additional_exp}
We further evaluate multiple CAAs in this section, and the experimental results on SVHN are also provided. In particular, we present the robust accuracy (RA) and their corresponding attack success rate (ASR). Again, the ASR is the percentage of the images that were initially classified correctly but were misclassified after being attacked; therefore, the lower ASR indicates the more robust model. 
In Sec. \ref{subsec:appendix_single_attack}, we especially show a single attack, which launches merely one attack from the attack pool. Notably, the $\ell_{\infty}$ (20-step) is regular PGD attack, and Auto-$\ell_{\infty}$ is an ensemble of four diverse attacks \cite{croce2020reliable}. Multiple attacks (including three, semantic, and full) are listed in Sec. \ref{subsec:appendix_multiple_attacks}. (For efficiency, we use $\ell_\infty$ (PGD) in multiple attack evaluation.)

\subsection{Single Attack}\label{subsec:appendix_single_attack}
\textbf{Results on CIFAR-10}
\input{preprint/assets/tables/cifar10/single_ra}
\input{preprint/assets/tables/cifar10/single_asr}

\clearpage
\textbf{Results on ImageNet}
\input{preprint/assets/tables/imagenet/single_ra}
\input{preprint/assets/tables/imagenet/single_asr}

\textbf{Results on SVHN}
\input{preprint/assets/tables/svhn/single_ra}
\input{preprint/assets/tables/svhn/single_asr}

\clearpage

\subsection{Multiple Attacks: Three attacks, Semantic attacks and Full attacks}\label{subsec:appendix_multiple_attacks}
We only provided the ASR of CIFAR-10 and ImageNet here; the RA can be found in Tables 1 and 2 of our paper. Again, the abbreviation used here is the same as in the paper.

\textbf{Results on CIFAR-10}
\vspace{-2mm}
\input{preprint/assets/tables/cifar10/multi_asr}

\textbf{Results on ImageNet}
\vspace{-2mm}
\input{preprint/assets/tables/imagenet/multi_asr}

\textbf{Results on SVHN}
\vspace{-2mm}
\input{preprint/assets/tables/svhn/multi_ra}
\vspace{-2mm}
\input{preprint/assets/tables/svhn/multi_asr}
\vspace{-2mm}

\section{Examples of Single Semantic Attacks at Different Levels}\label{sec:appendix_attack_levels}
Fig. \ref{fig:attack_levels} shows five single semantic attacks with corresponding perturbation levels. Each row represents the perturbed image of a corresponding attack $A_k$ with different perturbation values $\delta_{k}\in\epsilon_{k}$. 
\input{preprint/assets/figs/attack_exp/attack_exp}

\section{Additional Visualization of Adversarial Examples under Different CAA}
We present a series of adversarial examples from ImageNet generated using our proposed composite adversarial attacks (CAA). These attacks include single attacks, two attacks, three attacks, semantic attacks, and full attacks. To illustrate the effectiveness of our approach, we provide visualizations of the adversarial examples for each type of attack, arranged in several columns. Specifically, the left-most column of Figures \ref{fig:single_attack}, \ref{fig:two_attacks}, and \ref{fig:three_and_multiple} shows the original images. Every of the following two columns are the adversarial examples generated from one of the CAA attacks and their differences compared with the original images. Note that for visualization purposes only, all differences have been multiplied by three.

\input{preprint/assets/figs/attack_exp/demo/single_adv_exp}
\input{preprint/assets/figs/attack_exp/demo/two_adv_exp}
\input{preprint/assets/figs/attack_exp/demo/three_and_multiple}

%% file: preprint/assets/tables/epsilons.tex

\begin{wraptable}{r}{7cm}
\vspace{-3mm}
    \centering
\adjustbox{max width=7cm}{
    \centering
    \begin{tabular}{r|c|c}
    \toprule
    & \multicolumn{1}{c}{CIFAR-10, SVHN} & \multicolumn{1}{|l}{ImageNet} \\
    \midrule
    Hue, $\epsilon_{H}$ & \multicolumn{2}{c}{$-\pi \sim \pi$} \\
    \midrule
    Saturation, $\epsilon_{S}$ & \multicolumn{2}{c}{$0.7 \sim 1.3$} \\
    \midrule
    Rotation, $\epsilon_{R}$ & \multicolumn{2}{c}{$-10\degree \sim 10\degree$} \\
    \midrule
    Brightness, $\epsilon_{B}$ & \multicolumn{2}{c}{$-0.2 \sim 0.2$} \\
    \midrule
    Contrast, $\epsilon_{C}$ & \multicolumn{2}{c}{$0.7 \sim 1.3$} \\
    \midrule
    $\ell_{\infty}$, $\epsilon_L$ & 8/255 & 4/255 \\
    \bottomrule
    \end{tabular}}%
    \vspace{-2mm}
  \caption{Perturbation interval of each attack component}%
  \label{tab:attack_epsilons}%
\vspace{-3mm}
\end{wraptable}

%% file: preprint/assets/gat_algorithm.tex
\definecolor{comment}{RGB}{80, 80, 80}
\vspace{-3mm}
\begin{figure}[htbp]
  \centering
  \begin{minipage}{\linewidth}
\begin{algorithm}[H]
\setstretch{0.7}
    \SetKwData{IsAtt}{is\_attacked}
	\caption{Composite Adversarial Attack}
	\label{algo_composite}
	\KwIn{classifier $\mathcal{F}(\cdot)$, input $x$, label $y$, attack space $\pi=\{A_{1},\ldots,A_{n}\}$, attack order \textit{schedule}, scheduling iterations $M$, perturbation intervals $\{\epsilon_{k}\}_{k=1}^n$, Comp-PGD steps $T$}
	\KwOut{Composite adversarial examples $x_\text{c-adv}$}  
	\BlankLine
	\HiLi \# \textbf{Initialization}\\
	$\delta_{1}^{0},\ldots,\delta_{n}^{0} \gets $ initial perturbation\\
	\If{schedule == \textnormal{scheduled}}{
	$\mathcal{Z}^{0}$, $\pi_{0}\gets$ scheduling matrix and order assignment initialization}
        \Else{$\pi_{0} \gets$ initial order assignment (random / fixed)}
	\HiLi \# \textbf{Iteration of attack order scheduling}\\
	\For{$i\in\{1,\ldots,M\}$} {
	    \HiLi \# \textbf{Applying  attacks in order}\\
		\For{$k \in\{1,\ldots,n\}$}{
		    $A_{*} = A_{\pi_{i}(k)}$ \textnormal{, } $\delta_{*}^{0} = \delta_{\pi_{i}(k)}^{0}$ \textnormal{, }
		    $\epsilon_{*} = \epsilon_{\pi_{i}(k)}$ \\
		    $x_\text{c-adv}^k \gets A_{*}(x_\text{c-adv}^{k-1}; \delta_{*}^{0})$ \\
		    
		    \HiLi \# \textbf{Iteration of Comp-PGD}\\
            \For{$t\in\{1,\ldots,T\}$}{
		        \If{ $\mathcal{F}(x_\textnormal{c-adv}^k) \neq y$}{\label{algo:early_stopped}
		            \KwRet $x_\textnormal{c-adv}^k$ ~~~\textcolor{comment}{\# \textit{Early stop option}}
		        }
		        \Else{
		        $\delta_{*}^{t} = \textnormal{clip}_{\epsilon_{*}}(\cdot; x_\textnormal{c-adv}^{k}; \delta_{*}^{t-1})$ \textnormal{by Eq.} \ref{eqn:attack_pgd}\\
		        $x_\textnormal{c-adv}^{k} \gets A_{*}(x_\text{c-adv}^{k-1}; \delta_{*}^{t})$}
            }
		}
		\HiLi \# \textbf{Resetting the attack order}\\
		\If{schedule == \textnormal{random}}{
		    $\pi_{i+1} \gets$ \textnormal{Shuffle a new order}
		}
		\ElseIf{schedule == \textnormal{scheduled}}{
		\HiLi \# \textbf{Optimize scheduling order $Z$}\\
  
          $x_\text{surr} = \mathbf{z}_n^{\top}\mathbf{A}(\cdots(\mathbf{z}_{2}^{\top}\mathbf{A}(\mathbf{z}_{1}^{\top}\mathbf{A}(x))))$ ~~~\textcolor{comment}{\textit{\# Compute the surrogate composite adversarial example by Eq. \ref{eqn:surrogate_image}.}}\\
          
		 $Z^{t} = \mathcal{S}\big(\exp(Z^{t-1}+{\partial \mathcal{L}(\mathcal{F}(x_\text{surr}),y)}/{\partial Z^{t-1}})\big)$ ~~~\textcolor{comment}{\textit{\# Updating the scheduling matrix by Eq. \ref{eqn:dsm_update}}.}\\

		 $\pi_{i+1}(j) := \arg \max \mathbf{z}_{j} \text{, } \forall j \in \{1,\ldots,n\}$ ~~~\textcolor{comment}{\# \textit{Update the attack order assignment by Eq. \ref{eqn:hungarian}.}}
	}}
	\KwRet $x_\textnormal{c-adv}$
\end{algorithm}
  \end{minipage}
\end{figure}

%% file: preprint/assets/figs/loss_landscape/loss_traces.tex
\begin{figure*}[htbp]
     \centering
     \includegraphics[width=1\linewidth, trim=0.1cm 0.1cm 0.1cm 0.1cm, clip]{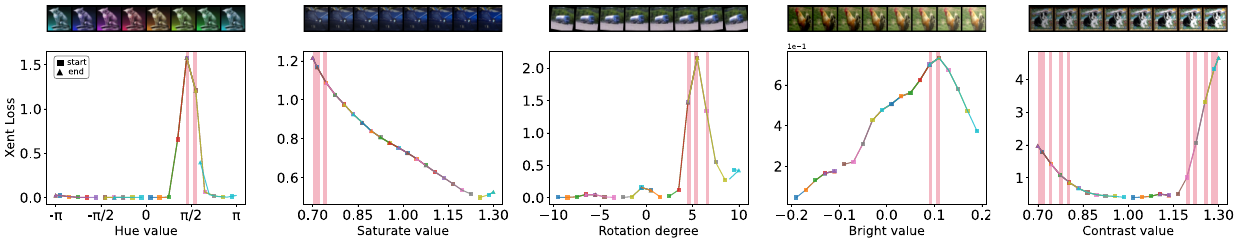}
     \caption{Component-wise PGD process of the single semantic attack. }
     \label{fig:appendix_loss_traces}
\end{figure*}

%% file: preprint/assets/tables/ablation1.tex
\begin{table}[!htb]
\centering
    \begin{tabular}{l|r|rrr|rr|rr}
    \toprule
    \multicolumn{1}{c|}{} & \multicolumn{1}{c|}{}  & \multicolumn{3}{c|}{Three attacks} & \multicolumn{2}{c|}{Semantic attacks}  & \multicolumn{2}{c}{Full attacks}\\
    Training  & Clean & \textit{CAA}$_{3a}$ & \textit{CAA}$_{3b}$ & \textit{CAA}$_{3c}$ & Rand. & Sched. & Rand. & Sched. \\
    \midrule
    Normal$^\dagger$ & 0.0   & 96.4  & 91.4  & 99.4  & 25.1  & 30.9  & 80.7  & 91.5  \\
    Madry$_{\infty}^{\dagger}$ & 0.0   & 43.2  & 54.7  & 57.9  & 37.6  & 46.3  & 57.7  & 72.0  \\
    PAT$_{self}^\dagger$ & 0.0   & 50.1  & 60.3  & 67.8  & 42.1  & 50.9  & 63.3  & 74.3  \\
    PAT$_{alex}^\dagger$ & 0.0   & 45.5  & 56.3  & 63.7  & 50.2  & 57.1  & 64.1  & 73.6  \\
    \textbf{GAT-f}$^\dagger$ & 0.0   & \textbf{50.9} & \textbf{50.8} & \textbf{63.0} & \textbf{7.4} & \textbf{9.0} & \textbf{39.7} & \textbf{56.2} \\
    \textbf{GAT-fs}$^\dagger$ & 0.0   & \textbf{47.3} & \textbf{50.0} & \textbf{52.7} & \textbf{7.7} & \textbf{9.3} & \textbf{40.3} & \textbf{52.5} \\
    \midrule
    Normal$^\ast$ & 0.0   & 96.8  & 89.9  & 99.6  & 32.3  & 38.8  & 83.9  & 87.6  \\
    Trades$_{\infty}^\ast$ & 0.0   & 40.7  & 52.7  & 66.2  & 48.5  & 58.3  & 64.2  & 74.3  \\
    FAT$_{\infty}^\ast$ & 0.0   & 42.1  & 57.6  & 64.9  & 47.9  & 59.6  & 65.8  & 76.6  \\
    AWP$_{\infty}^\ast$ & 0.0   & 37.2  & 50.1  & 62.5  & 50.0  & 58.3  & 64.6  & 77.1  \\
    \textbf{GAT-f}$^\ast$ & 0.0   & \textbf{47.0} & \textbf{49.0} & \textbf{52.8} & \textbf{6.8} & \textbf{8.8} & \textbf{41.1} & \textbf{54.2} \\
    \textbf{GAT-fs}$^\ast$ & 0.0   & \textbf{46.3} & \textbf{48.8} & \textbf{52.9} & \textbf{6.9} & \textbf{8.7} & \textbf{40.0} & \textbf{53.9} \\
    \bottomrule
    \end{tabular}
    \caption{Attack success rate (\%) of using Ensemble-PGD to perform CAA on CIFAR-10.}
    \label{tab:cifar10_ablation}%
\end{table}

%% file: preprint/assets/tables/ablation2.tex
\begin{table}[htbp]
\setlength\tabcolsep{4.15pt}
\centering
    \begin{tabular}{l|r|rrr|rr|rr}
    \toprule
    \multicolumn{1}{c|}{} & \multicolumn{1}{c|}{}  & \multicolumn{3}{c|}{Three attacks} & \multicolumn{2}{c|}{Semantic attacks}  & \multicolumn{2}{c}{Full attacks}\\
    Training  & Clean & \textit{CAA}$_{3a}$ & \textit{CAA}$_{3b}$ & \textit{CAA}$_{3c}$ & Rand. & Sched. & Rand. & Sched. \\
    \midrule
RSP$^\ast$ & 84.9  & 17.9 {\footnotesize $\pm$ 0.7} & 11.6 {\footnotesize $\pm$ 0.9} & 5.9 {\footnotesize $\pm$ 0.6} & 22.8 {\footnotesize $\pm$ 0.2} & 12.6 {\footnotesize $\pm$ 0.0} & 6.7 {\footnotesize $\pm$ 0.3} & 1.6 {\footnotesize $\pm$ 0.2} \\
RSP-N$^\ast$ & 88.1  & 18.8 {\footnotesize $\pm$ 0.4} & 11.9 {\footnotesize $\pm$ 0.3} & 8.8 {\footnotesize $\pm$ 0.5} & 39.4 {\footnotesize $\pm$ 0.5} & 28.5 {\footnotesize $\pm$ 0.3} & 10.2 {\footnotesize $\pm$ 0.2} & 3.9 {\footnotesize $\pm$ 0.3} \\
RSP-T$^\ast$ & 85.4  & 38.3 {\footnotesize $\pm$ 0.2} & 28.4 {\footnotesize $\pm$ 0.2} & 21.2 {\footnotesize $\pm$ 0.5} & 54.6 {\footnotesize $\pm$ 0.3} & 47.5 {\footnotesize $\pm$ 0.7} & 20.4 {\footnotesize $\pm$ 0.7} & 9.8 {\footnotesize $\pm$ 1.0} \\
\midrule
\textbf{GAT-f}$^\ast$ & \textbf{83.4} & \textbf{40.2 {\footnotesize $\pmb{\pm}$ 0.1}} & \textbf{34.0 {\footnotesize $\pmb{\pm}$ 0.1}} & \textbf{30.7 {\footnotesize $\pmb{\pm}$ 0.4}} & \textbf{71.6 {\footnotesize $\pmb{\pm}$ 0.1}} & \textbf{67.8 {\footnotesize $\pmb{\pm}$ 0.2}} & \textbf{31.2 {\footnotesize $\pmb{\pm}$ 0.4}} & \textbf{20.1 {\footnotesize $\pmb{\pm}$ 0.3}} \\
\textbf{GAT-fs}$^\ast$ & \textbf{83.2} & \textbf{43.5 {\footnotesize $\pmb{\pm}$ 0.1}} & \textbf{36.3 {\footnotesize $\pmb{\pm}$ 0.1}} & \textbf{32.9 {\footnotesize $\pmb{\pm}$ 0.4}} & \textbf{70.5 {\footnotesize $\pmb{\pm}$ 0.1}} & \textbf{66.7 {\footnotesize $\pmb{\pm}$ 0.3}} & \textbf{32.2 {\footnotesize $\pmb{\pm}$ 0.7}} & \textbf{21.9 {\footnotesize $\pmb{\pm}$ 0.7}} \\
    \bottomrule
    \multicolumn{9}{l}{\footnotesize \textit{Note.} RSP$^\ast$: AT from scratch; RSP-N$^\ast$: AT, fine-tuned on Normal$^\ast$; RSP-T$^\ast$: AT, fine-tuned on Trades$^\ast_\infty$} \\
    \end{tabular}
\vspace{-2mm}
    \caption{Robust accuracy (\%) of models trained with simulated CAA samples.}
    \label{tab:appendix_rsp_cifar10_ra}%
\vspace{-8mm}
\end{table} 

%% file: preprint/assets/tables/imagenet/two_asr.tex
\begin{table}[h]
  \centering
    \begin{tabular}{l|rrrrr}
    \toprule
     & \multicolumn{5}{c}{2 attacks (Semantic $\to\ell_\infty$)} \\
    Training & {Hue $\to\ell_\infty$} & {Saturation $\to\ell_\infty$} & {Rotation $\to\ell_\infty$} & {Brightness $\to\ell_\infty$} & {Contrast $\to\ell_\infty$} \\
    
    \midrule
    Normal$^\dagger$ & 100.0 {\footnotesize $\pm$ 0.0} & 100.0 {\footnotesize $\pm$ 0.0} & 100.0 {\footnotesize $\pm$ 0.0} & 100.0 {\footnotesize $\pm$ 0.0} & 100.0 {\footnotesize $\pm$ 0.0} \\
    Madry$_{\infty}^{\dagger}$ & 72.3 {\footnotesize $\pm$ 2.0} & 52.4 {\footnotesize $\pm$ 0.7} & 60.6 {\footnotesize $\pm$ 0.3} & 56.5 {\footnotesize $\pm$ 0.5} & 58.2 {\footnotesize $\pm$ 0.7} \\
    Fast-AT$_{\infty}^\dagger$ & 76.7 {\footnotesize $\pm$ 1.8} & 56.7 {\footnotesize $\pm$ 0.6} & 66.6 {\footnotesize $\pm$ 0.1} & 62.4 {\footnotesize $\pm$ 0.7} & 63.9 {\footnotesize $\pm$ 0.7} \\
    \textbf{GAT-f}$^\dagger$ & 60.9 {\footnotesize $\pm$ 2.4} & 59.6 {\footnotesize $\pm$ 0.8} & 60.8 {\footnotesize $\pm$ 0.6} & 59.7 {\footnotesize $\pm$ 0.6} & 64.2 {\footnotesize $\pm$ 0.5} \\

    \midrule
     & \multicolumn{5}{c}{2 attacks ($\ell_\infty\to$ \text{Semantic})} \\
     Training & {$\ell_\infty\to$ Hue} & {$\ell_\infty\to$ Saturation} & {$\ell_\infty\to$ Rotation} & {$\ell_\infty\to$ Brightness} & {$\ell_\infty\to$ Contrast} \\
    \midrule
    Normal$^\dagger$ & 100.0 {\footnotesize $\pm$ 0.0} & 100.0 {\footnotesize $\pm$ 0.0} & 99.9 {\footnotesize $\pm$ 0.0} & 100.0 {\footnotesize $\pm$ 0.0} & 100.0 {\footnotesize $\pm$ 0.0} \\
    Madry$_{\infty}^{\dagger}$ & 64.6 {\footnotesize $\pm$ 0.5} \textbf{(7.7$\downarrow$)} & 49.1 {\footnotesize $\pm$ 0.5} \textbf{(3.3$\downarrow$)} & 52.5 {\footnotesize $\pm$ 0.1} \textbf{(8.0$\downarrow$)} & 50.9 {\footnotesize $\pm$ 0.4} \textbf{(5.7$\downarrow$)} & 48.8 {\footnotesize $\pm$ 0.5} \textbf{(9.4$\downarrow$)} \\
    Fast-AT$_{\infty}^\dagger$ & 71.5 {\footnotesize $\pm$ 0.6} \textbf{(5.2$\downarrow$)} & 54.1 {\footnotesize $\pm$ 0.6} \textbf{(2.6$\downarrow$)} & 60.0 {\footnotesize $\pm$ 0.0} \textbf{(6.6$\downarrow$)} & 58.6 {\footnotesize $\pm$ 0.5} \textbf{(3.8$\downarrow$)} & 56.5 {\footnotesize $\pm$ 0.4} \textbf{(7.4$\downarrow$)} \\
    \textbf{GAT-f}$^\dagger$ & 60.8 {\footnotesize $\pm$ 0.5} \textbf{(0.1$\downarrow$)} & 58.2 {\footnotesize $\pm$ 0.8} \textbf{(1.4$\downarrow$)} & 56.5 {\footnotesize $\pm$ 0.9} \textbf{(4.2$\downarrow$)} & 57.2 {\footnotesize $\pm$ 0.8} \textbf{(2.5$\downarrow$)} & 57.6 {\footnotesize $\pm$ 0.8} \textbf{(6.7$\downarrow$)} \\

    \bottomrule
    \end{tabular}%
  \caption{Attack success rate of two attacks with two order settings on ImageNet. The value in the parenthesis is the reduced value compare with another order settings.}%
  \label{tab:appendix_imagenet_two_asr}%
\end{table}

%% file: preprint/assets/figs/order-effect/order-effect.tex
\begin{figure}[htbp]
  \subfloat[Attack order 1: \underline{$\ell_\infty$} $\to$ Hue $\to$ Saturation $\to$ Rotation $\to$ Brightness $\to$ Contrast]{
	\begin{minipage}[c]{\linewidth}
	   \centering
	   \includegraphics[width=.8\linewidth]{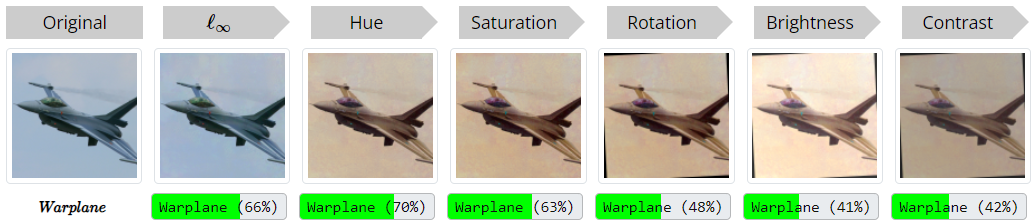}
	\end{minipage}}
 \hfill
  \subfloat[Attack order 2: Hue $\to$ Saturation $\to$ Rotation  $\to$ \underline{$\ell_\infty$} $\to$ Brightness$\to$ Contrast]{
	\begin{minipage}[c]{\linewidth}
	   \centering
	   \includegraphics[width=.8\linewidth]{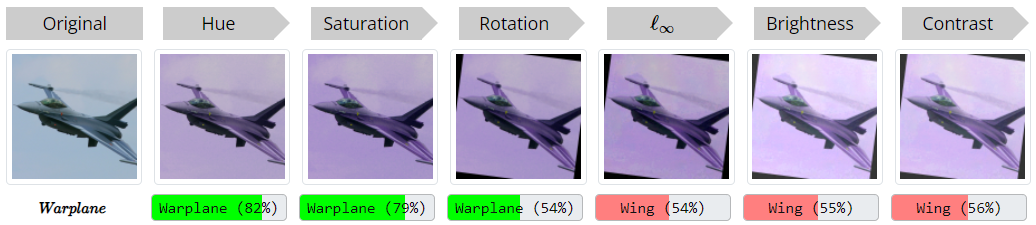}
	\end{minipage}}
  \caption{Warplane example. (a) The model can correctly predict objects under a given attack order. (b) The model can easily be fooled when adding $\ell_{\infty}$ perturbations after Hue, Saturation, and Rotation attacks. (\textit{Note: Attack parameters are optimized by Comp-PGD algorithm.})}
  \label{fig:appendix_order_effect_1}
  \vspace{-12mm}
\end{figure}

%% file: preprint/assets/tables/cifar10/single_ra.tex
\begin{table}[H]
  \centering
    \begin{tabular}{l|r|rrrrrr|r} 
    \toprule
    \multicolumn{1}{c|}{} &        & \multicolumn{6}{c|}{Single attack} & \multicolumn{1}{c}{Auto attack} \\
    Training & \multicolumn{1}{c|}{Clean} & Hue   & Saturation & Rotation & Brightness & Contrast & $\ell_{\infty}$ (20-step) & Auto-$\ell_{\infty}$ \\
    \midrule
Normal$^\dagger$ & 95.2  & 81.8 {\footnotesize $\pm$ 0.0} & 94.0 {\footnotesize $\pm$ 0.0} & 88.1 {\footnotesize $\pm$ 0.1} & 92.1 {\footnotesize $\pm$ 0.1} & 93.7 {\footnotesize $\pm$ 0.1} & 0.0 {\footnotesize $\pm$ 0.0} & 0.0 {\footnotesize $\pm$ 0.0} \\
Madry$_{\infty}^{\dagger}$ & 87.0  & 70.8 {\footnotesize $\pm$ 0.0} & 84.8 {\footnotesize $\pm$ 0.0} & 79.5 {\footnotesize $\pm$ 0.1} & 77.0 {\footnotesize $\pm$ 0.1} & 79.9 {\footnotesize $\pm$ 0.1} & 53.5 {\footnotesize $\pm$ 0.0} & 49.2 {\footnotesize $\pm$ 0.0} \\
PAT$_{self}^\dagger$ & 82.4  & 64.3 {\footnotesize $\pm$ 0.2} & 79.8 {\footnotesize $\pm$ 0.0} & 74.1 {\footnotesize $\pm$ 0.1} & 72.5 {\footnotesize $\pm$ 0.1} & 78.0 {\footnotesize $\pm$ 0.1} & 41.2 {\footnotesize $\pm$ 0.0} & 30.2 {\footnotesize $\pm$ 0.0} \\
PAT$_{alex}^\dagger$ & 71.6  & 53.2 {\footnotesize $\pm$ 0.2} & 68.9 {\footnotesize $\pm$ 0.0} & 63.8 {\footnotesize $\pm$ 0.1} & 60.6 {\footnotesize $\pm$ 0.1} & 65.2 {\footnotesize $\pm$ 0.0} & 41.9 {\footnotesize $\pm$ 0.0} & 28.8 {\footnotesize $\pm$ 0.0} \\
\textbf{GAT-f}$^\dagger$ & 82.3  & 81.2 {\footnotesize $\pm$ 0.5} & 80.8 {\footnotesize $\pm$ 0.1} & 78.3 {\footnotesize $\pm$ 0.5} & 80.1 {\footnotesize $\pm$ 0.1} & 79.7 {\footnotesize $\pm$ 0.1} & 42.7 {\footnotesize $\pm$ 0.0} & 38.7 {\footnotesize $\pm$ 0.0} \\
\textbf{GAT-fs}$^\dagger$ & 82.1  & 80.6 {\footnotesize $\pm$ 0.0} & 80.8 {\footnotesize $\pm$ 0.0} & 78.0 {\footnotesize $\pm$ 0.2} & 80.4 {\footnotesize $\pm$ 0.1} & 79.5 {\footnotesize $\pm$ 0.1} & 46.6 {\footnotesize $\pm$ 0.0} & 41.9 {\footnotesize $\pm$ 0.0} \\
\midrule
Normal$^\ast$ & 94.0  & 75.8 {\footnotesize $\pm$ 0.2} & 92.3 {\footnotesize $\pm$ 0.0} & 87.4 {\footnotesize $\pm$ 0.2} & 89.1 {\footnotesize $\pm$ 0.0} & 91.3 {\footnotesize $\pm$ 0.1} & 0.0 {\footnotesize $\pm$ 0.0} & 0.0 {\footnotesize $\pm$ 0.0} \\
Trades$_{\infty}^\ast$ & 84.9  & 65.7 {\footnotesize $\pm$ 0.2} & 82.7 {\footnotesize $\pm$ 0.0} & 77.5 {\footnotesize $\pm$ 0.1} & 69.7 {\footnotesize $\pm$ 0.3} & 70.7 {\footnotesize $\pm$ 0.1} & 55.8 {\footnotesize $\pm$ 0.0} & 52.5 {\footnotesize $\pm$ 0.0} \\
FAT$_{\infty}^\ast$ & 88.1  & 69.0 {\footnotesize $\pm$ 0.1} & 85.4 {\footnotesize $\pm$ 0.0} & 77.0 {\footnotesize $\pm$ 0.1} & 73.1 {\footnotesize $\pm$ 0.4} & 76.5 {\footnotesize $\pm$ 0.1} & 54.7 {\footnotesize $\pm$ 0.0} & 51.5 {\footnotesize $\pm$ 0.0} \\
AWP$_{\infty}^\ast$ & 85.4  & 67.5 {\footnotesize $\pm$ 0.1} & 83.0 {\footnotesize $\pm$ 0.0} & 77.0 {\footnotesize $\pm$ 0.2} & 68.3 {\footnotesize $\pm$ 0.1} & 70.8 {\footnotesize $\pm$ 0.0} & 59.4 {\footnotesize $\pm$ 0.0} & 56.2 {\footnotesize $\pm$ 0.0} \\
\textbf{GAT-f}$^\ast$ & 83.4  & 82.3 {\footnotesize $\pm$ 0.6} & 81.8 {\footnotesize $\pm$ 0.0} & 79.5 {\footnotesize $\pm$ 0.4} & 81.7 {\footnotesize $\pm$ 0.0} & 81.0 {\footnotesize $\pm$ 0.1} & 43.6 {\footnotesize $\pm$ 0.0} & 40.0 {\footnotesize $\pm$ 0.0} \\
\textbf{GAT-fs}$^\ast$ & 83.2  & 81.5 {\footnotesize $\pm$ 0.1} & 81.7 {\footnotesize $\pm$ 0.0} & 78.8 {\footnotesize $\pm$ 0.0} & 81.2 {\footnotesize $\pm$ 0.0} & 80.7 {\footnotesize $\pm$ 0.1} & 47.2 {\footnotesize $\pm$ 0.0} & 42.2 {\footnotesize $\pm$ 0.0} \\

    \bottomrule
    \end{tabular}%
  \caption{Robust accuracy of single attack, which is one of semantic attacks, on CIFAR-10}%
  \label{tab:appendix_cifar10_sin_ra}%
\end{table}%

%% file: preprint/assets/tables/cifar10/single_asr.tex
\begin{table}[H]
  \centering
    \begin{tabular}{l|r|rrrrrr|r} 
    \toprule
    \multicolumn{1}{c|}{} &        & \multicolumn{6}{c|}{Single attack} & \multicolumn{1}{c}{Auto attack} \\
    Training & \multicolumn{1}{c|}{Clean} & Hue   & Saturation & Rotation & Brightness & Contrast & $\ell_{\infty}$ (20-step) & Auto-$\ell_{\infty}$ \\
    \midrule
Normal$^\dagger$ & 0.0   & 14.4 {\footnotesize $\pm$ 0.0} & 1.4 {\footnotesize $\pm$ 0.0} & 8.0 {\footnotesize $\pm$ 0.2} & 3.3 {\footnotesize $\pm$ 0.1} & 1.6 {\footnotesize $\pm$ 0.1} & 100.0 {\footnotesize $\pm$ 0.0} & 100.0 {\footnotesize $\pm$ 0.0} \\
Madry$_{\infty}^{\dagger}$ & 0.0   & 19.3 {\footnotesize $\pm$ 0.0} & 2.6 {\footnotesize $\pm$ 0.0} & 9.3 {\footnotesize $\pm$ 0.1} & 11.7 {\footnotesize $\pm$ 0.2} & 8.4 {\footnotesize $\pm$ 0.0} & 38.6 {\footnotesize $\pm$ 0.0} & 43.4 {\footnotesize $\pm$ 0.0} \\
PAT$_{self}^\dagger$ & 0.0   & 23.0 {\footnotesize $\pm$ 0.2} & 3.3 {\footnotesize $\pm$ 0.0} & 11.8 {\footnotesize $\pm$ 0.2} & 12.6 {\footnotesize $\pm$ 0.1} & 5.7 {\footnotesize $\pm$ 0.1} & 50.0 {\footnotesize $\pm$ 0.0} & 63.4 {\footnotesize $\pm$ 0.0} \\
PAT$_{alex}^\dagger$ & 0.0   & 27.7 {\footnotesize $\pm$ 0.2} & 4.2 {\footnotesize $\pm$ 0.1} & 13.0 {\footnotesize $\pm$ 0.1} & 17.2 {\footnotesize $\pm$ 0.2} & 10.0 {\footnotesize $\pm$ 0.0} & 41.4 {\footnotesize $\pm$ 0.0} & 59.8 {\footnotesize $\pm$ 0.0} \\
\textbf{GAT-f}$^\dagger$ & 0.0   & 1.6 {\footnotesize $\pm$ 0.5} & 1.9 {\footnotesize $\pm$ 0.1} & 5.7 {\footnotesize $\pm$ 0.5} & 2.8 {\footnotesize $\pm$ 0.1} & 3.3 {\footnotesize $\pm$ 0.0} & 48.2 {\footnotesize $\pm$ 0.0} & 53.0 {\footnotesize $\pm$ 0.0} \\
\textbf{GAT-fs}$^\dagger$ & 0.0   & 2.0 {\footnotesize $\pm$ 0.0} & 1.6 {\footnotesize $\pm$ 0.0} & 5.8 {\footnotesize $\pm$ 0.2} & 2.2 {\footnotesize $\pm$ 0.1} & 3.3 {\footnotesize $\pm$ 0.1} & 43.2 {\footnotesize $\pm$ 0.0} & 49.0 {\footnotesize $\pm$ 0.0} \\
    \midrule
\cmidrule{1-1}Normal$^\ast$ & 0.0   & 19.7 {\footnotesize $\pm$ 0.2} & 1.8 {\footnotesize $\pm$ 0.0} & 7.6 {\footnotesize $\pm$ 0.2} & 5.3 {\footnotesize $\pm$ 0.1} & 2.9 {\footnotesize $\pm$ 0.1} & 100.0 {\footnotesize $\pm$ 0.0} & 100.0 {\footnotesize $\pm$ 0.0} \\
Trades$_{\infty}^\ast$ & 0.0   & 23.0 {\footnotesize $\pm$ 0.2} & 2.6 {\footnotesize $\pm$ 0.0} & 9.5 {\footnotesize $\pm$ 0.1} & 18.2 {\footnotesize $\pm$ 0.3} & 16.9 {\footnotesize $\pm$ 0.1} & 34.3 {\footnotesize $\pm$ 0.0} & 38.2 {\footnotesize $\pm$ 0.0} \\
FAT$_{\infty}^\ast$ & 0.0   & 22.2 {\footnotesize $\pm$ 0.1} & 3.1 {\footnotesize $\pm$ 0.0} & 13.3 {\footnotesize $\pm$ 0.1} & 17.3 {\footnotesize $\pm$ 0.5} & 13.4 {\footnotesize $\pm$ 0.1} & 37.9 {\footnotesize $\pm$ 0.0} & 41.5 {\footnotesize $\pm$ 0.0} \\
AWP$_{\infty}^\ast$ & 0.0   & 21.5 {\footnotesize $\pm$ 0.2} & 2.8 {\footnotesize $\pm$ 0.0} & 10.5 {\footnotesize $\pm$ 0.2} & 20.3 {\footnotesize $\pm$ 0.1} & 17.3 {\footnotesize $\pm$ 0.0} & 30.4 {\footnotesize $\pm$ 0.0} & 34.2 {\footnotesize $\pm$ 0.0} \\
\textbf{GAT-f}$^\ast$ & 0.0   & 1.7 {\footnotesize $\pm$ 0.5} & 2.0 {\footnotesize $\pm$ 0.0} & 5.6 {\footnotesize $\pm$ 0.3} & 2.3 {\footnotesize $\pm$ 0.0} & 3.1 {\footnotesize $\pm$ 0.0} & 47.7 {\footnotesize $\pm$ 0.0} & 52.0 {\footnotesize $\pm$ 0.0} \\
\textbf{GAT-fs}$^\ast$ & 0.0   & 2.3 {\footnotesize $\pm$ 0.1} & 1.9 {\footnotesize $\pm$ 0.0} & 5.9 {\footnotesize $\pm$ 0.0} & 2.6 {\footnotesize $\pm$ 0.0} & 3.2 {\footnotesize $\pm$ 0.0} & 43.3 {\footnotesize $\pm$ 0.0} & 49.2 {\footnotesize $\pm$ 0.0} \\

    \bottomrule
    \end{tabular}%
  \caption{Attack success rate of single attack on CIFAR-10.}%
  \label{tab:appendix_cifar10_sin_asr}%
\end{table}%

%% file: preprint/assets/tables/imagenet/single_ra.tex
\begin{table}[H]
  \centering
    \begin{tabular}{l|r|rrrrrr|r} 
    \toprule
    \multicolumn{1}{c|}{} &        & \multicolumn{6}{c|}{Single attack} & \multicolumn{1}{c}{Auto attack} \\
    Training & \multicolumn{1}{c|}{Clean} & Hue   & Saturation & Rotation & Brightness & Contrast & $\ell_{\infty}$ (20-step) & Auto-$\ell_{\infty}$ \\
    \midrule
Normal$^\dagger$ & 76.1  & 50.9 {\footnotesize $\pm$ 0.2} & 72.5 {\footnotesize $\pm$ 0.1} & 68.2 {\footnotesize $\pm$ 0.6} & 69.2 {\footnotesize $\pm$ 0.3} & 71.8 {\footnotesize $\pm$ 0.2} & 0.0 {\footnotesize $\pm$ 0.0} & 0.0 {\footnotesize $\pm$ 0.0} \\
Madry$_{\infty}^{\dagger}$ & 62.4  & 38.1 {\footnotesize $\pm$ 0.4} & 58.4 {\footnotesize $\pm$ 0.0} & 51.0 {\footnotesize $\pm$ 0.7} & 53.6 {\footnotesize $\pm$ 0.1} & 55.9 {\footnotesize $\pm$ 0.0} & 33.5 {\footnotesize $\pm$ 0.0} & 28.9 {\footnotesize $\pm$ 0.0} \\
Fast-AT$_{\infty}^\dagger$ & 53.8  & 27.8 {\footnotesize $\pm$ 0.2} & 48.0 {\footnotesize $\pm$ 0.0} & 38.6 {\footnotesize $\pm$ 0.6} & 42.0 {\footnotesize $\pm$ 0.1} & 44.0 {\footnotesize $\pm$ 0.0} & 27.5 {\footnotesize $\pm$ 0.0} & 24.7 {\footnotesize $\pm$ 0.0} \\
\textbf{GAT-f}$^\dagger$ & 60.0  & 51.0 {\footnotesize $\pm$ 2.5} & 58.1 {\footnotesize $\pm$ 0.0} & 56.5 {\footnotesize $\pm$ 0.3} & 57.7 {\footnotesize $\pm$ 0.1} & 58.1 {\footnotesize $\pm$ 0.1} & 25.2 {\footnotesize $\pm$ 0.0} & 20.9 {\footnotesize $\pm$ 0.0} \\
    \bottomrule
    \end{tabular}%
  \caption{Robust accuracy of single attack, which is one of semantic attacks, on ImageNet.}%
  \label{tab:appendix_imagenet_sin_ra}%
\end{table}%

%% file: preprint/assets/tables/imagenet/single_asr.tex
\begin{table}[H]
  \centering
    \begin{tabular}{l|r|rrrrrr|r} 
    \toprule
    \multicolumn{1}{c|}{} &        & \multicolumn{6}{c|}{Single attack} & \multicolumn{1}{c}{Auto attack} \\
    Training & \multicolumn{1}{c|}{Clean} & Hue   & Saturation & Rotation & Brightness & Contrast & $\ell_{\infty}$ (20-step) & Auto-$\ell_{\infty}$ \\
    \midrule
Normal$^\dagger$ & 0.0   & 34.3 {\footnotesize $\pm$ 0.2} & 4.9 {\footnotesize $\pm$ 0.1} & 12.2 {\footnotesize $\pm$ 0.6} & 9.5 {\footnotesize $\pm$ 0.3} & 5.9 {\footnotesize $\pm$ 0.2} & 100.0 {\footnotesize $\pm$ 0.0} & 100.0 {\footnotesize $\pm$ 0.0} \\
Madry$_{\infty}^{\dagger}$ & 0.0   & 39.9 {\footnotesize $\pm$ 0.6} & 6.7 {\footnotesize $\pm$ 0.0} & 19.9 {\footnotesize $\pm$ 0.9} & 14.6 {\footnotesize $\pm$ 0.1} & 10.8 {\footnotesize $\pm$ 0.1} & 46.3 {\footnotesize $\pm$ 0.0} & 53.6 {\footnotesize $\pm$ 0.0} \\
Fast-AT$_{\infty}^\dagger$ & 0.0   & 49.3 {\footnotesize $\pm$ 0.3} & 11.1 {\footnotesize $\pm$ 0.0} & 29.8 {\footnotesize $\pm$ 0.9} & 22.8 {\footnotesize $\pm$ 0.1} & 19.0 {\footnotesize $\pm$ 0.1} & 48.9 {\footnotesize $\pm$ 0.0} & 54.1 {\footnotesize $\pm$ 0.0} \\
\textbf{GAT-f}$^\dagger$ & 0.0   & 17.7 {\footnotesize $\pm$ 3.2} & 3.3 {\footnotesize $\pm$ 0.0} & 6.8 {\footnotesize $\pm$ 0.3} & 3.9 {\footnotesize $\pm$ 0.1} & 3.3 {\footnotesize $\pm$ 0.1} & 57.9 {\footnotesize $\pm$ 0.0} & 65.1 {\footnotesize $\pm$ 0.0} \\
    \bottomrule
    \end{tabular}%
  \caption{Attack success rate of single attack, on ImageNet}%
  \label{tab:appendix_imagenet_sin_asr}%
\end{table}%

%% file: preprint/assets/tables/svhn/single_ra.tex
\begin{table}[H]
  \centering
    \begin{tabular}{l|r|rrrrrr|r} 
    \toprule
    \multicolumn{1}{c|}{} &        & \multicolumn{6}{c|}{Single attack} & \multicolumn{1}{c}{Auto attack} \\
    Training & \multicolumn{1}{c|}{Clean} & Hue   & Saturation & Rotation & Brightness & Contrast & $\ell_{\infty}$ (20-step) & Auto-$\ell_{\infty}$ \\
    \midrule
Normal$^\ast$ & 95.4  & 93.3 {\footnotesize $\pm$ 0.0} & 94.7 {\footnotesize $\pm$ 0.0} & 89.7 {\footnotesize $\pm$ 0.1} & 92.2 {\footnotesize $\pm$ 0.0} & 93.7 {\footnotesize $\pm$ 0.0} & 0.5 {\footnotesize $\pm$ 0.0} & 0.0 {\footnotesize $\pm$ 0.0} \\
Trades$_{\infty}^{\ast}$ & 90.3  & 87.3 {\footnotesize $\pm$ 0.1} & 89.2 {\footnotesize $\pm$ 0.0} & 81.4 {\footnotesize $\pm$ 0.0} & 77.2 {\footnotesize $\pm$ 0.1} & 83.3 {\footnotesize $\pm$ 0.1} & 53.3 {\footnotesize $\pm$ 0.0} & 44.2 {\footnotesize $\pm$ 0.0} \\
\textbf{GAT-f}$^\ast$ & 93.4  & 92.5 {\footnotesize $\pm$ 0.0} & 93.0 {\footnotesize $\pm$ 0.0} & 91.2 {\footnotesize $\pm$ 0.0} & 92.1 {\footnotesize $\pm$ 0.1} & 92.1 {\footnotesize $\pm$ 0.0} & 51.2 {\footnotesize $\pm$ 0.0} & 36.9 {\footnotesize $\pm$ 0.0} \\
\textbf{GAT-fs}$^\ast$ & 93.6  & 92.8 {\footnotesize $\pm$ 0.0} & 93.1 {\footnotesize $\pm$ 0.0} & 91.7 {\footnotesize $\pm$ 0.0} & 92.5 {\footnotesize $\pm$ 0.0} & 92.3 {\footnotesize $\pm$ 0.0} & 54.1 {\footnotesize $\pm$ 0.0} & 38.2 {\footnotesize $\pm$ 0.0} \\
    \bottomrule
    \end{tabular}%
  \caption{Robust accuracy of single attack, which is one of semantic attacks, on SVHN.}%
  \label{tab:appendix_svhn_sin_ra}%
\end{table}%

%% file: preprint/assets/tables/svhn/single_asr.tex
\begin{table}[H]
  \centering
    \begin{tabular}{l|r|rrrrrr|r} 
    \toprule
    \multicolumn{1}{c|}{} &        & \multicolumn{6}{c|}{Single attack} & \multicolumn{1}{c}{Auto attack} \\
    Training & \multicolumn{1}{c|}{Clean} & Hue   & Saturation & Rotation & Brightness & Contrast & $\ell_{\infty}$ (20-step) & Auto-$\ell_{\infty}$ \\
    \midrule
Normal$^\ast$ & 0.0   & 2.4 {\footnotesize $\pm$ 0.0} & 0.7 {\footnotesize $\pm$ 0.0} & 6.2 {\footnotesize $\pm$ 0.1} & 3.4 {\footnotesize $\pm$ 0.0} & 1.8 {\footnotesize $\pm$ 0.0} & 99.5 {\footnotesize $\pm$ 0.0} & 100.0 {\footnotesize $\pm$ 0.0} \\
Trades$_{\infty}^{\ast}$ & 0.0   & 4.0 {\footnotesize $\pm$ 0.1} & 1.3 {\footnotesize $\pm$ 0.0} & 10.0 {\footnotesize $\pm$ 0.0} & 14.9 {\footnotesize $\pm$ 0.1} & 8.0 {\footnotesize $\pm$ 0.0} & 41.0 {\footnotesize $\pm$ 0.0} & 51.0 {\footnotesize $\pm$ 0.0} \\
\textbf{GAT-f}$^\ast$ & 0.0   & 1.1 {\footnotesize $\pm$ 0.0} & 0.5 {\footnotesize $\pm$ 0.0} & 2.5 {\footnotesize $\pm$ 0.0} & 1.5 {\footnotesize $\pm$ 0.1} & 1.5 {\footnotesize $\pm$ 0.0} & 45.2 {\footnotesize $\pm$ 0.1} & 60.5 {\footnotesize $\pm$ 0.0} \\
\textbf{GAT-fs}$^\ast$ & 0.0   & 1.0 {\footnotesize $\pm$ 0.0} & 0.6 {\footnotesize $\pm$ 0.0} & 2.2 {\footnotesize $\pm$ 0.1} & 1.2 {\footnotesize $\pm$ 0.0} & 1.4 {\footnotesize $\pm$ 0.0} & 42.2 {\footnotesize $\pm$ 0.0} & 59.1 {\footnotesize $\pm$ 0.0} \\

    \bottomrule
    \end{tabular}%
  \caption{Attack success rate of single attack, on SVHN}%
  \label{tab:appendix_svhn_sin_asr}%
\end{table}%

%% file: preprint/assets/tables/cifar10/multi_asr.tex
\begin{table}[h]
  \centering
    \begin{tabular}{l|r|rrr|rr|rr}
    \toprule
    \multicolumn{1}{c|}{} & \multicolumn{1}{c|}{}  & \multicolumn{3}{c|}{Three attacks} & \multicolumn{2}{c|}{Semantic attacks}  & \multicolumn{2}{c}{Full attacks}\\
    Training  & Clean & \textit{CAA}$_{3a}$ & \textit{CAA}$_{3b}$ & \textit{CAA}$_{3c}$ & Rand. & Sched. & Rand. & Sched. \\
    \midrule
Normal$^\dagger$ & 0.0   & 100.0 {\footnotesize $\pm$ 0.0} & 100.0 {\footnotesize $\pm$ 0.0} & 100.0 {\footnotesize $\pm$ 0.0} & 37.4 {\footnotesize $\pm$ 0.3} & 53.6 {\footnotesize $\pm$ 0.5} & 100.0 {\footnotesize $\pm$ 0.0} & 100.0 {\footnotesize $\pm$ 0.0} \\
Madry$_{\infty}^{\dagger}$ & 0.0   & 64.6 {\footnotesize $\pm$ 0.2} & 78.4 {\footnotesize $\pm$ 0.6} & 78.0 {\footnotesize $\pm$ 0.3} & 63.8 {\footnotesize $\pm$ 0.2} & 75.5 {\footnotesize $\pm$ 0.3} & 87.6 {\footnotesize $\pm$ 0.2} & 95.8 {\footnotesize $\pm$ 0.2} \\
PAT$_{self}^\dagger$ & 0.0   & 74.7 {\footnotesize $\pm$ 0.2} & 85.6 {\footnotesize $\pm$ 0.5} & 78.3 {\footnotesize $\pm$ 0.4} & 65.0 {\footnotesize $\pm$ 0.3} & 78.8 {\footnotesize $\pm$ 0.4} & 88.9 {\footnotesize $\pm$ 0.4} & 96.9 {\footnotesize $\pm$ 0.3} \\
PAT$_{alex}^\dagger$ & 0.0   & 71.1 {\footnotesize $\pm$ 0.5} & 82.5 {\footnotesize $\pm$ 0.2} & 77.0 {\footnotesize $\pm$ 0.6} & 67.6 {\footnotesize $\pm$ 0.4} & 82.9 {\footnotesize $\pm$ 0.6} & 85.7 {\footnotesize $\pm$ 0.1} & 96.5 {\footnotesize $\pm$ 0.3} \\
\textbf{GAT-f}$^\dagger$ & 0.0   & 51.6 {\footnotesize $\pm$ 0.2} & 59.6 {\footnotesize $\pm$ 0.1} & 64.9 {\footnotesize $\pm$ 0.3} & 15.2 {\footnotesize $\pm$ 0.1} & 19.8 {\footnotesize $\pm$ 0.1} & 63.5 {\footnotesize $\pm$ 0.5} & 77.1 {\footnotesize $\pm$ 0.4} \\
\textbf{GAT-fs}$^\dagger$ & 0.0   & 47.0 {\footnotesize $\pm$ 0.1} & 55.4 {\footnotesize $\pm$ 0.2} & 60.5 {\footnotesize $\pm$ 0.2} & 15.0 {\footnotesize $\pm$ 0.2} & 18.8 {\footnotesize $\pm$ 0.1} & 60.7 {\footnotesize $\pm$ 1.0} & 73.5 {\footnotesize $\pm$ 0.4} \\
\midrule
Normal$^\ast$ & 0.0   & 100.0 {\footnotesize $\pm$ 0.0} & 100.0 {\footnotesize $\pm$ 0.0} & 100.0 {\footnotesize $\pm$ 0.0} & 51.1 {\footnotesize $\pm$ 0.4} & 68.3 {\footnotesize $\pm$ 0.6} & 100.0 {\footnotesize $\pm$ 0.0} & 100.0 {\footnotesize $\pm$ 0.0} \\
Trades$_{\infty}^\ast$ & 0.0   & 64.7 {\footnotesize $\pm$ 0.4} & 76.7 {\footnotesize $\pm$ 0.7} & 88.1 {\footnotesize $\pm$ 0.6} & 80.4 {\footnotesize $\pm$ 0.3} & 90.5 {\footnotesize $\pm$ 0.5} & 93.2 {\footnotesize $\pm$ 0.4} & 98.2 {\footnotesize $\pm$ 0.2} \\
FAT$_{\infty}^\ast$ & 0.0   & 66.2 {\footnotesize $\pm$ 0.4} & 80.6 {\footnotesize $\pm$ 0.5} & 85.5 {\footnotesize $\pm$ 0.7} & 78.8 {\footnotesize $\pm$ 0.2} & 88.9 {\footnotesize $\pm$ 0.5} & 93.0 {\footnotesize $\pm$ 0.2} & 98.3 {\footnotesize $\pm$ 0.1} \\
AWP$_{\infty}^\ast$ & 0.0   & 59.9 {\footnotesize $\pm$ 0.2} & 72.8 {\footnotesize $\pm$ 0.3} & 87.0 {\footnotesize $\pm$ 0.4} & 81.8 {\footnotesize $\pm$ 0.2} & 90.7 {\footnotesize $\pm$ 0.2} & 93.1 {\footnotesize $\pm$ 0.1} & 98.0 {\footnotesize $\pm$ 0.2} \\
\textbf{GAT-f}$^\ast$ & 0.0   & 51.8 {\footnotesize $\pm$ 0.1} & 59.2 {\footnotesize $\pm$ 0.1} & 63.2 {\footnotesize $\pm$ 0.5} & 14.2 {\footnotesize $\pm$ 0.1} & 18.7 {\footnotesize $\pm$ 0.2} & 62.6 {\footnotesize $\pm$ 0.4} & 75.9 {\footnotesize $\pm$ 0.4} \\
\textbf{GAT-fs}$^\ast$ & 0.0   & 47.8 {\footnotesize $\pm$ 0.2} & 56.3 {\footnotesize $\pm$ 0.1} & 60.5 {\footnotesize $\pm$ 0.4} & 15.4 {\footnotesize $\pm$ 0.1} & 19.8 {\footnotesize $\pm$ 0.4} & 61.4 {\footnotesize $\pm$ 0.9} & 73.7 {\footnotesize $\pm$ 0.8} \\
    \bottomrule
    \end{tabular}
  \caption{Attack success rate of composite semantic attacks and composite full attacks on CIFAR-10.}%
  \label{tab:appendix_cifar10_multi_asr}%
\end{table}%

%% file: preprint/assets/tables/imagenet/multi_asr.tex
\begin{table}[H]
  \centering
    \begin{tabular}{l|r|rrr|rr|rr}
    \toprule
    \multicolumn{1}{c|}{} & \multicolumn{1}{c|}{}  & \multicolumn{3}{c|}{Three attacks} & \multicolumn{2}{c|}{Semantic attacks}  & \multicolumn{2}{c}{Full attacks}\\
    Training  & Clean & \textit{CAA}$_{3a}$ & \textit{CAA}$_{3b}$ & \textit{CAA}$_{3c}$ & Rand. & Sched. & Rand. & Sched. \\
    \midrule
Normal$^\dagger$ & 0.0   & 100.0 {\footnotesize $\pm$ 0.0} & 100.0 {\footnotesize $\pm$ 0.0} & 100.0 {\footnotesize $\pm$ 0.0} & 59.1 {\footnotesize $\pm$ 0.5} & 72.9 {\footnotesize $\pm$ 1.3} & 100.0 {\footnotesize $\pm$ 0.0} & 100.0 {\footnotesize $\pm$ 0.0} \\
Madry$_{\infty}^{\dagger}$ & 0.0   & 77.7 {\footnotesize $\pm$ 0.6} & 85.2 {\footnotesize $\pm$ 0.4} & 74.0 {\footnotesize $\pm$ 1.3} & 77.6 {\footnotesize $\pm$ 0.1} & 85.6 {\footnotesize $\pm$ 0.1} & 88.7 {\footnotesize $\pm$ 0.2} & 95.5 {\footnotesize $\pm$ 0.4} \\
Fast-AT$_{\infty}^\dagger$ & 0.0   & 82.3 {\footnotesize $\pm$ 0.5} & 89.8 {\footnotesize $\pm$ 0.2} & 78.9 {\footnotesize $\pm$ 1.5} & 88.3 {\footnotesize $\pm$ 0.1} & 93.3 {\footnotesize $\pm$ 0.1} & 94.3 {\footnotesize $\pm$ 0.2} & 98.1 {\footnotesize $\pm$ 0.2} \\
\textbf{GAT-f}$^\dagger$ & 0.0   & 67.9 {\footnotesize $\pm$ 1.7} & 68.4 {\footnotesize $\pm$ 2.3} & 69.2 {\footnotesize $\pm$ 0.7} & 27.8 {\footnotesize $\pm$ 3.0} & 34.6 {\footnotesize $\pm$ 3.3} & 69.1 {\footnotesize $\pm$ 0.9} & 80.3 {\footnotesize $\pm$ 0.2} \\
    \bottomrule
    \end{tabular}
  \caption{Attack success rate of composite semantic attacks and composite full attacks on ImageNet.}%
  \label{tab:appendix_imagenet_multi_asr}%
\end{table}%

%% file: preprint/assets/tables/svhn/multi_ra.tex
\begin{table}[H]
  \centering
    \begin{tabular}{l|r|rrr|rr|rr}
    \toprule
    \multicolumn{1}{c|}{} & \multicolumn{1}{c|}{}  & \multicolumn{3}{c|}{Three attacks} & \multicolumn{2}{c|}{Semantic attacks}  & \multicolumn{2}{c}{Full attacks}\\
    Training  & Clean & \textit{CAA}$_{3a}$ & \textit{CAA}$_{3b}$ & \textit{CAA}$_{3c}$ & Rand. & Sched. & Rand. & Sched. \\
    \midrule
Normal$^\ast$ & 95.4  & 0.4 {\footnotesize $\pm$ 0.0} & 0.2 {\footnotesize $\pm$ 0.0} & 0.4 {\footnotesize $\pm$ 0.1} & 78.5 {\footnotesize $\pm$ 0.2} & 68.7 {\footnotesize $\pm$ 0.6} & 0.5 {\footnotesize $\pm$ 0.0} & 0.2 {\footnotesize $\pm$ 0.1} \\
Trades$_{\infty}^{\ast}$ & 90.3  & 43.6 {\footnotesize $\pm$ 0.1} & 32.1 {\footnotesize $\pm$ 0.3} & 21.2 {\footnotesize $\pm$ 0.7} & 47.3 {\footnotesize $\pm$ 0.2} & 34.7 {\footnotesize $\pm$ 0.5} & 22.6 {\footnotesize $\pm$ 0.5} & 10.6 {\footnotesize $\pm$ 0.4} \\
\textbf{GAT-f}$^\ast$ & 93.4  & 47.0 {\footnotesize $\pm$ 0.1} & 42.8 {\footnotesize $\pm$ 0.3} & 34.4 {\footnotesize $\pm$ 0.5} & 85.5 {\footnotesize $\pm$ 0.1} & 82.8 {\footnotesize $\pm$ 0.2} & 37.1 {\footnotesize $\pm$ 0.2} & 26.8 {\footnotesize $\pm$ 0.6} \\
\textbf{GAT-fs}$^\ast$ & 93.6  & 48.7 {\footnotesize $\pm$ 0.1} & 45.2 {\footnotesize $\pm$ 0.3} & 35.6 {\footnotesize $\pm$ 0.5} & 86.6 {\footnotesize $\pm$ 0.1} & 83.7 {\footnotesize $\pm$ 0.2} & 39.0 {\footnotesize $\pm$ 0.4} & 28.2 {\footnotesize $\pm$ 0.3} \\
    \bottomrule
    \end{tabular}
  \caption{Robust accuracy of composite semantic attacks and composite full attacks on SVHN.}%
  \label{tab:appendix_svhn_multi_ra}%
\end{table}%

%% file: preprint/assets/tables/svhn/multi_asr.tex
\begin{table}[H]
  \centering
    \begin{tabular}{l|r|rrr|rr|rr}
    \toprule
    \multicolumn{1}{c|}{} & \multicolumn{1}{c|}{}  & \multicolumn{3}{c|}{Three attacks} & \multicolumn{2}{c|}{Semantic attacks}  & \multicolumn{2}{c}{Full attacks}\\
    Training  & Clean & \textit{CAA}$_{3a}$ & \textit{CAA}$_{3b}$ & \textit{CAA}$_{3c}$ & Rand. & Sched. & Rand. & Sched. \\
    \midrule
Normal$^\ast$ & 0.0   & 99.5 {\footnotesize $\pm$ 0.0} & 99.8 {\footnotesize $\pm$ 0.0} & 99.6 {\footnotesize $\pm$ 0.1} & 17.8 {\footnotesize $\pm$ 0.3} & 28.0 {\footnotesize $\pm$ 0.6} & 99.5 {\footnotesize $\pm$ 0.0} & 99.8 {\footnotesize $\pm$ 0.1} \\
Trades$_{\infty}^{\ast}$ & 0.0   & 51.7 {\footnotesize $\pm$ 0.1} & 64.4 {\footnotesize $\pm$ 0.3} & 76.5 {\footnotesize $\pm$ 0.8} & 47.6 {\footnotesize $\pm$ 0.3} & 61.6 {\footnotesize $\pm$ 0.6} & 75.0 {\footnotesize $\pm$ 0.5} & 88.2 {\footnotesize $\pm$ 0.5} \\
\textbf{GAT-f}$^\ast$ & 0.0   & 49.7 {\footnotesize $\pm$ 0.1} & 54.2 {\footnotesize $\pm$ 0.3} & 63.2 {\footnotesize $\pm$ 0.5} & 8.5 {\footnotesize $\pm$ 0.1} & 11.4 {\footnotesize $\pm$ 0.2} & 60.3 {\footnotesize $\pm$ 0.2} & 71.3 {\footnotesize $\pm$ 0.7} \\
\textbf{GAT-fs}$^\ast$ & 0.0   & 48.0 {\footnotesize $\pm$ 0.1} & 51.7 {\footnotesize $\pm$ 0.3} & 62.0 {\footnotesize $\pm$ 0.6} & 7.5 {\footnotesize $\pm$ 0.1} & 10.6 {\footnotesize $\pm$ 0.2} & 58.3 {\footnotesize $\pm$ 0.4} & 69.8 {\footnotesize $\pm$ 0.3} \\
    \bottomrule
    \end{tabular}
  \caption{Attack success rate of composite semantic attacks and composite full attacks on SVHN.}%
  \label{tab:appendix_svhn_multi_asr}%
\end{table}%

%% file: preprint/assets/figs/attack_exp/attack_exp.tex
\begin{figure}[!h]
    \begin{center}
    \resizebox{8.2cm}{!}{
\begingroup
    \begin{tabular}{@{\hskip -0.05in}r@{\hskip 0.05in}l}
    \vspace{-0.2cm}\textbf{Hue} & \adjustimage{height=2cm,valign=m}{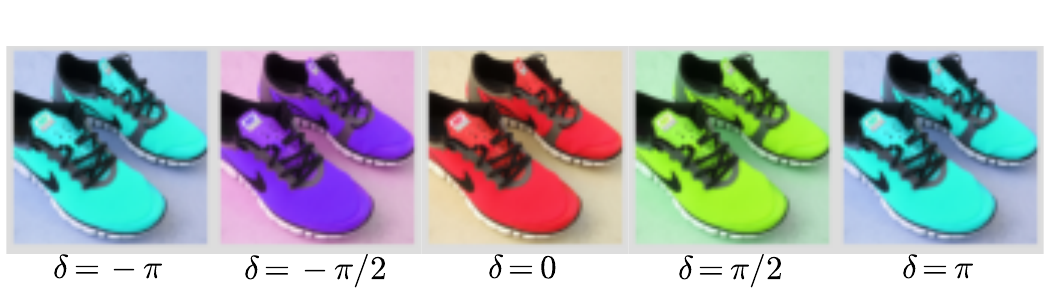} \\
    \vspace{-0.2cm} \textbf{Saturation} & \adjustimage{height=2cm,valign=m}{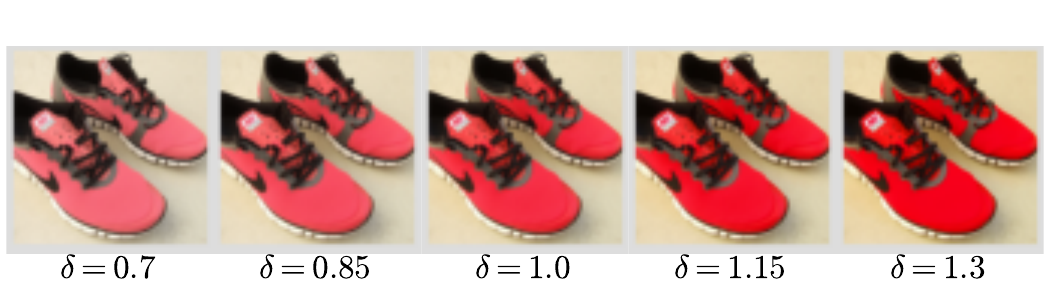} \\
    \vspace{-0.2cm} \textbf{Rotation} &
    \adjustimage{height=2cm,valign=m}{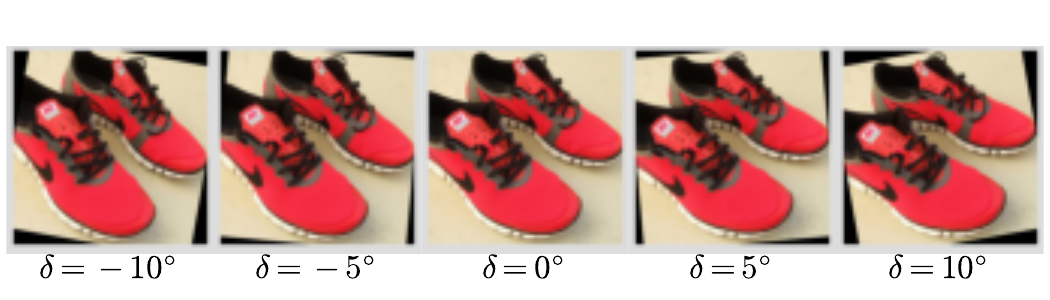} \\
    \vspace{-0.2cm} \textbf{Brightness} & \adjustimage{height=2cm,valign=m}{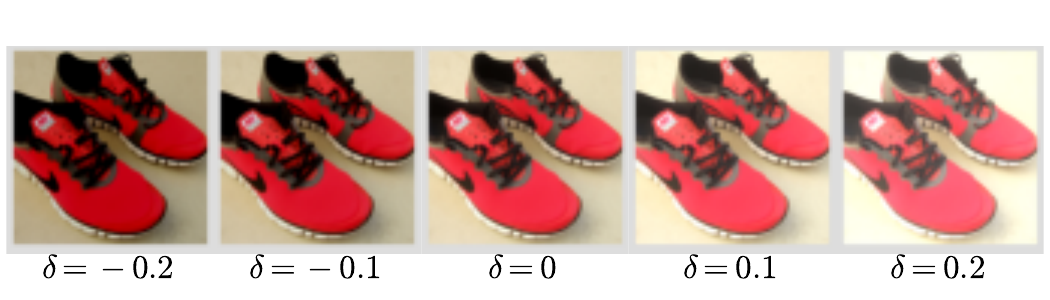} \\
    \vspace{-0.2cm} \textbf{Contrast} & \adjustimage{height=2cm,valign=m}{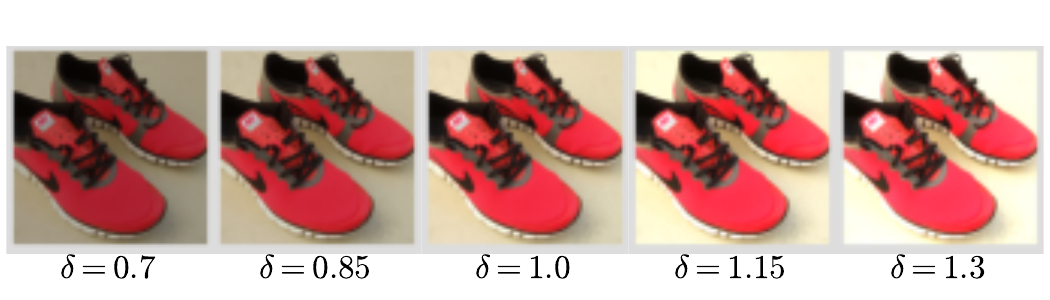} \\
    \end{tabular}
\endgroup
    }
    \end{center}
\caption{Single Semantic Attack Examples. Clean image was placed at the center of each row.}\label{fig:attack_levels}
    \vspace{-0.5cm}
\end{figure}%

%% file: preprint/assets/figs/attack_exp/demo/single_adv_exp.tex
\begin{figure}[ht]
    \centering
    \includegraphics[width=.9\textwidth]{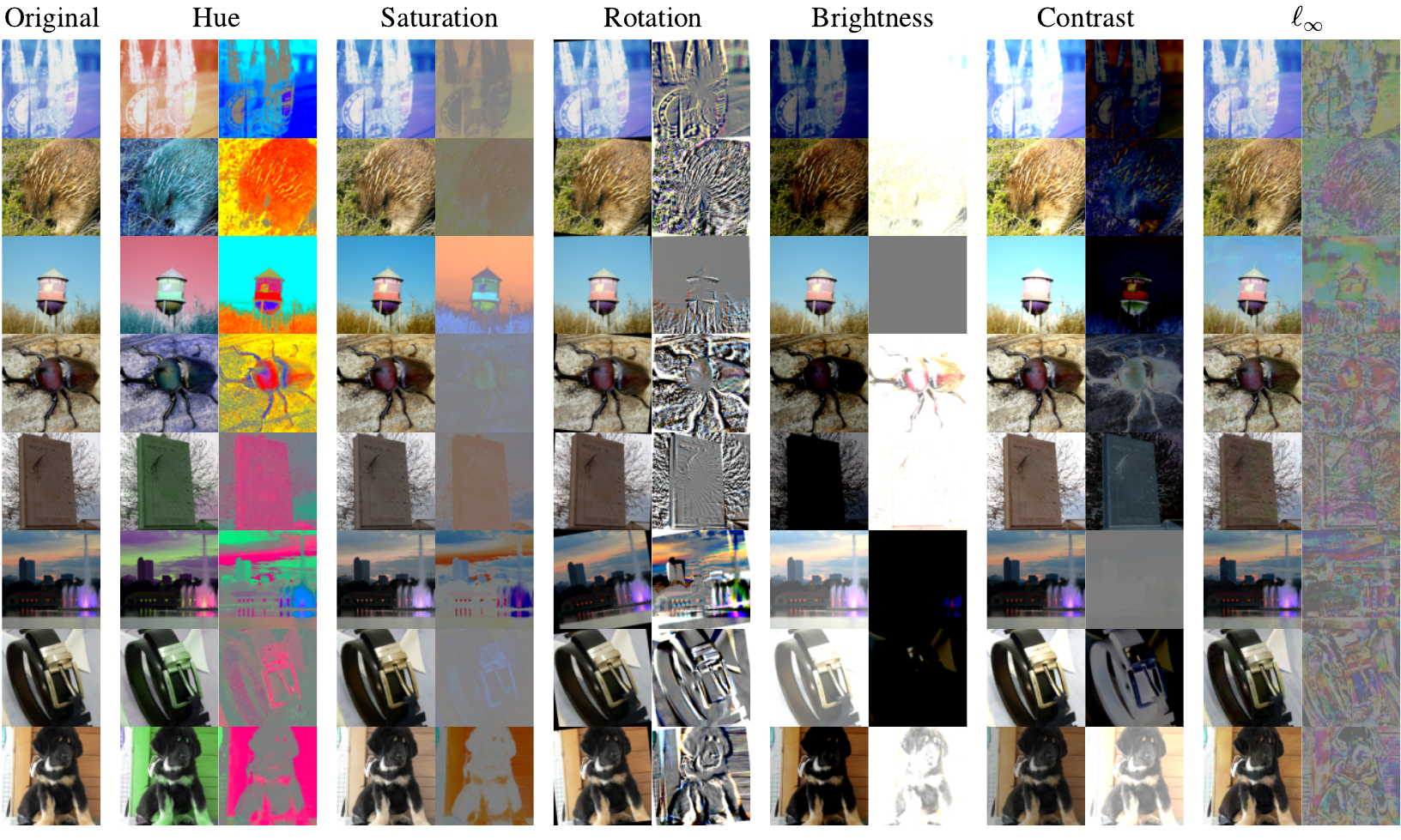}
    \caption{Adversarial examples generated under \textbf{single semantic} attacks or $\ell_{\infty}$ attack.}
    \label{fig:single_attack}
\end{figure}

%% file: preprint/assets/figs/attack_exp/demo/two_adv_exp.tex
\begin{figure}[H]
    \centering
    \includegraphics[width=.9\textwidth]{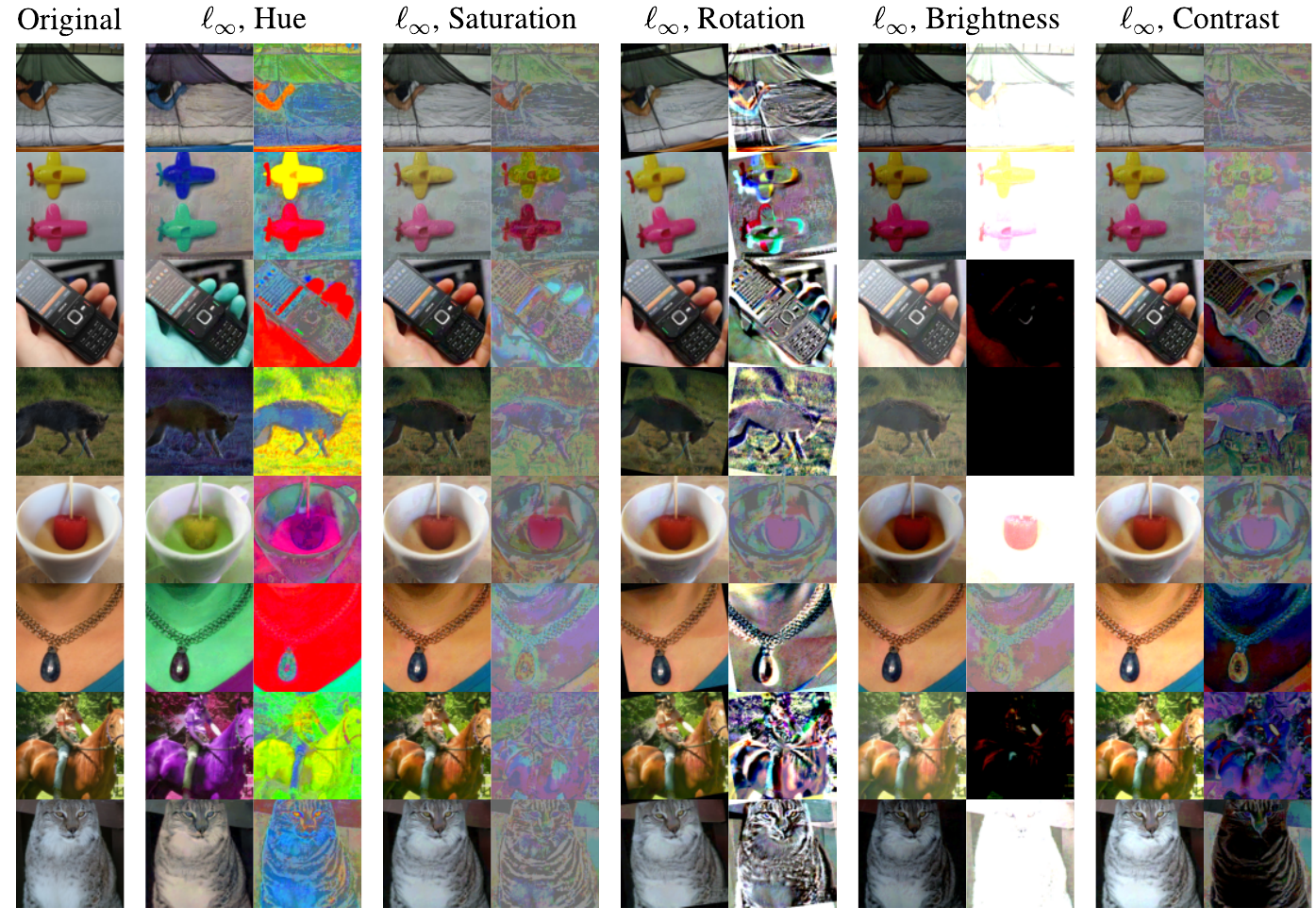}
    \caption{Composite adversarial examples generated under \textbf{two attacks} (composed of one semantic attack and the $\ell_{\infty}$ attack).}
    \label{fig:two_attacks}
\end{figure}

%% file: preprint/assets/figs/attack_exp/demo/three_and_multiple.tex
\begin{figure}[H]
    \centering
    \includegraphics[width=.9\textwidth]{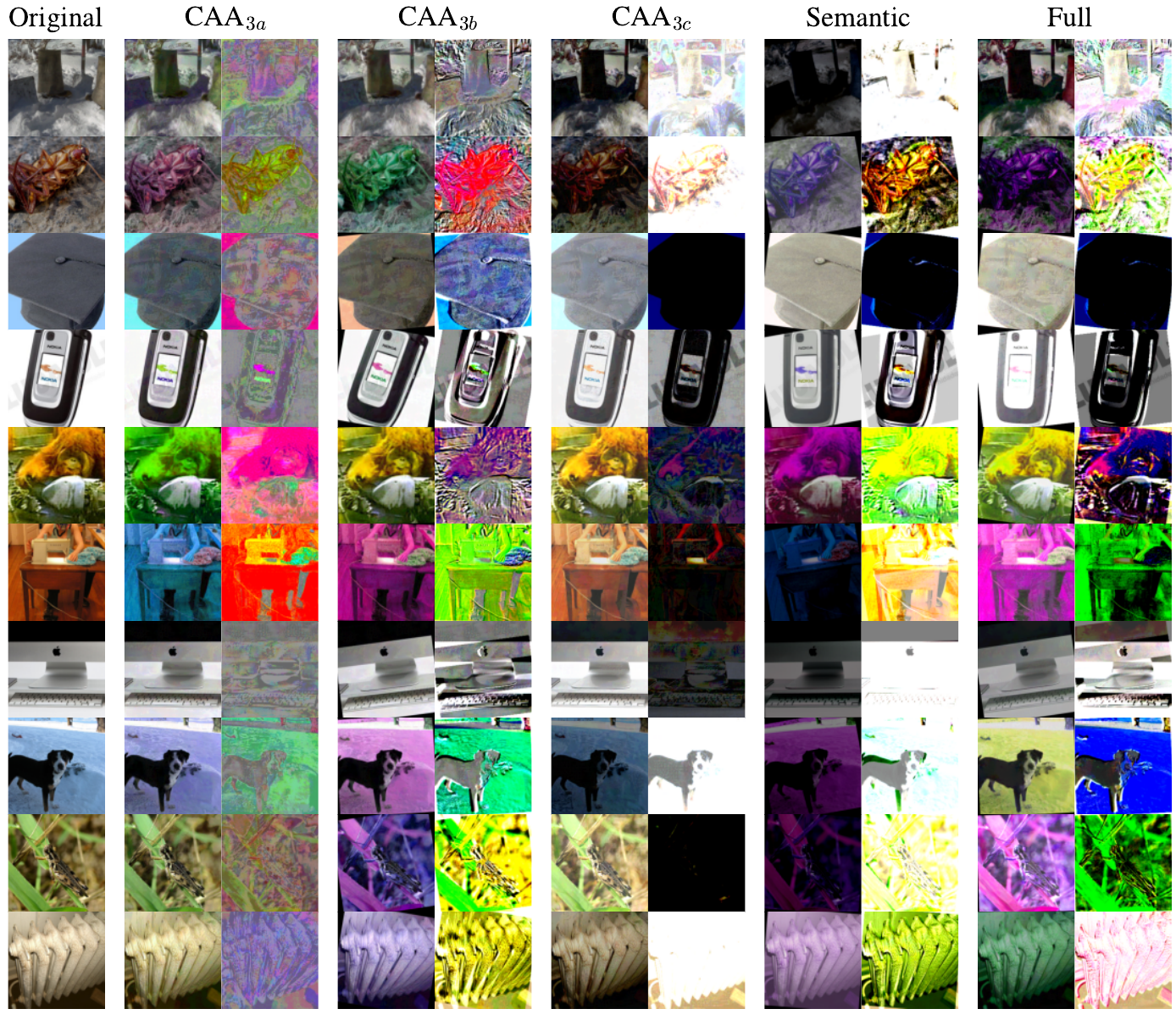}
    \caption{Composite adversarial examples generated under \textbf{three attacks and other multiple attacks}. \textit{CAA}$_{3a}$: (Hue, Saturation, $\ell_{\infty}$), \textit{CAA}$_{3b}$: (Hue, Rotation, $\ell_{\infty}$), \textit{CAA}$_{3c}$: (Brightness, Contrast, $\ell_{\infty}$), Semantic: (Hue, Saturation, Rotation, Brightness, and Contrast), and Full: \textit{$\ell_{\infty}+$ all semantic attacks}.}
    \label{fig:three_and_multiple}
\end{figure}